\documentclass{article}

\usepackage{arxiv}

\usepackage[utf8]{inputenc} % allow utf-8 input
\usepackage[T1]{fontenc}    % use 8-bit T1 fonts
\usepackage{hyperref}       % hyperlinks
\usepackage{url}            % simple URL typesetting
\usepackage{booktabs}       % professional-quality tables
\usepackage{amsfonts}       % blackboard math symbols
\usepackage{nicefrac}       % compact symbols for 1/2, etc.
\usepackage{microtype}      % microtypography
\usepackage{lipsum}		% Can be removed after putting your text content
\usepackage{graphicx}
\usepackage[numbers]{natbib}
\usepackage{doi}
\usepackage{wrapfig}
\usepackage{amsmath}
\usepackage{multirow}
\usepackage{booktabs}
\usepackage{graphicx}
\usepackage{adjustbox}
\usepackage[absolute,overlay]{textpos}
\usepackage{caption}
\usepackage[table]{xcolor}
\definecolor{lightgreen}{RGB}{230,245,230}
\usepackage{float}

\usepackage{xcolor}
\definecolor{darkgreen}{RGB}{0, 200, 0}

\title{\textcolor{darkgreen}{Vi}sion Meets \textcolor{darkgreen}{Wi}Fi \\
Physics-Grounded Estimation of Volumetric Mechanical Properties
}

\author{
Ali Bahri$^{1}$ \quad
Hongliang Li$^{1}$ \quad
Soufiane Lamghari$^{1}$ \quad
Jie Chuai$^{2}$ \quad
Zhitang Chen$^{2}$
\\[1mm]
$^{1}$Huawei Noah's Ark Lab, Canada \\
$^{2}$Huawei Noah's Ark Lab, Hong Kong SAR, China
}

\renewcommand{\shorttitle}{\textit{arXiv} Template}

\hypersetup{
pdftitle={A template for the arxiv style},
pdfsubject={q-bio.NC, q-bio.QM},
pdfauthor={Ali Bahri, Hongliang Li, Soufiane Lamghari, Jie Chuai, Zhitang Chen},
pdfkeywords={First keyword, Second keyword, More},
}

\renewcommand{\undertitle}{\vspace{-4mm}A Preprint}
\date{}
\begin{document}
\maketitle

\begin{center}
    \includegraphics[width=\textwidth]{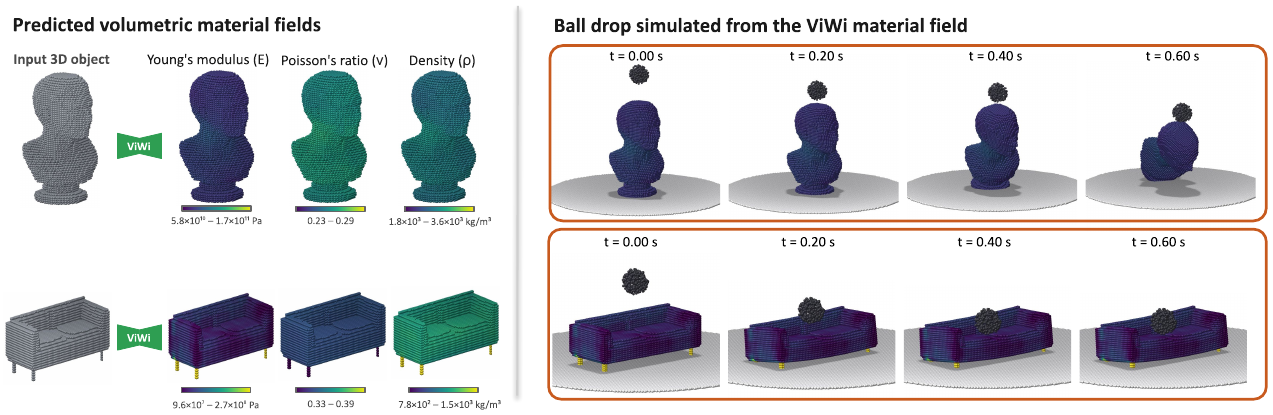}
    \vspace{-4pt}
    \captionof{figure}{
    \textbf{From volumetric mechanical-property estimation to physics simulation.}
    ViWi predicts voxel-level Young's modulus, Poisson's ratio, and density from
    the input 3D object. The predicted material fields are then used to simulate
    the object's physical response to a ball impact.
    }
    \label{fig:viwi_banner}
\end{center}

\begin{abstract}
Estimating volumetric mechanical properties, including Young's modulus, Poisson's
ratio, and density at each voxel, is intrinsically ambiguous from vision alone, as
visually similar objects may have substantially different material compositions and
physical behavior. Existing approaches predict these properties independently across
voxels, overlooking the piecewise-constant material structure of real objects and
producing noisy or inconsistent estimates for voxels that share the same material,
while lacking an explicit mechanism to resolve visual ambiguity. We introduce
\textbf{ViWi} (\textbf{Vi}sion Meets \textbf{Wi}Fi), an object-centric framework
for volumetric mechanical-property estimation. ViWi represents each object using a
compact set of material slots that aggregate evidence from voxels with a shared
material identity and produce coherent slot-level property predictions. To complement
visual appearance, ViWi incorporates a compact RF descriptor generated through
WiFi-band electromagnetic simulation using permittivity and conductivity. The RF
descriptor conditions the material slots with global composition cues that may be
unavailable from images, while visual features preserve voxel-level spatial
localization. On GVM, ViWi improves over the prior state of the art on four of six
per-voxel metrics, while its vision-only variant improves all reported mass-estimation
metrics on ABO-500. These results demonstrate that combining object-centric material
structure with complementary RF evidence enables more accurate and physically coherent
volumetric property estimation beyond what is possible from visual appearance alone.
\end{abstract}

% keywords can be removed

\section{Introduction}
\label{sec:introduction}

\begin{wrapfigure}[24]{r}{0.39\textwidth}
    \centering
    \vspace{-45pt}
    \includegraphics[width=\linewidth]{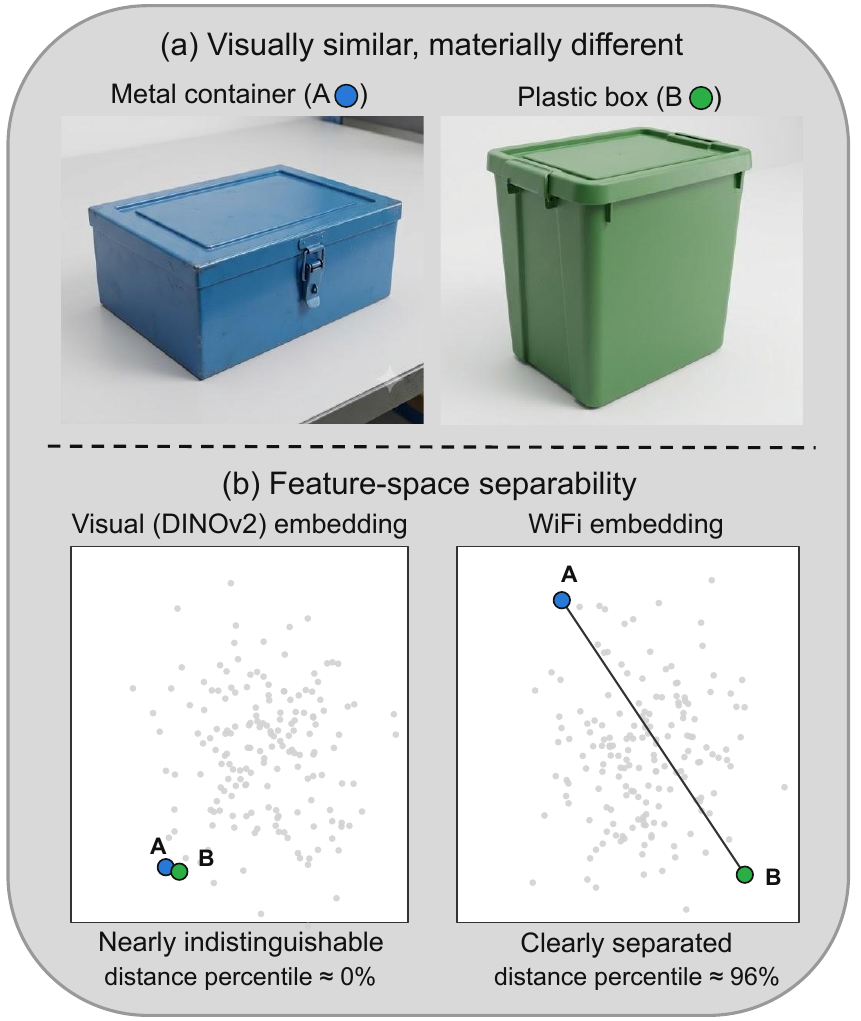}
    \caption{
    % \textbf{Visual ambiguity and RF separability.}
    % \textbf{(a)} Visually similar objects with different material compositions.
    % \textbf{(b)} The pair overlaps in the visual embedding but is clearly separated
    % in the RF embedding.
    \textbf{Visual ambiguity and RF separability.}
    \textbf{(a)} Illustrative renderings of a metal container (A) and a plastic
    box (B). \textbf{(b)} Embeddings of the actual assets: the pair ranks in the
    closest $0.4\%$ of pairs visually, but the most distant $96\%$ in RF.
    }
    \label{fig:motivation}
\end{wrapfigure}

Understanding how objects deform, resist external forces, and distribute mass is
essential for physically grounded perception. Applications such as robotic
manipulation, physics-based simulation, digital twins, augmented and virtual
reality, and embodied intelligence require more than an object's geometry or
semantic identity; they require spatially resolved estimates of its mechanical
properties. In particular, Young's modulus \(E\), Poisson's ratio \(\nu\), and
density \(\rho\) determine an object's stiffness, lateral deformation, and mass
distribution, respectively. Accurate volumetric estimates of these quantities
enable downstream systems to anticipate physical behavior, select safe interaction
strategies, and construct simulation-ready object representations. However,
assigning mechanical properties to 3D assets remains largely manual, requiring
material annotations, expert knowledge, or direct physical measurements that are
difficult to obtain at scale.

% \hspace{20 pt}

Recent work estimates physical and mechanical properties directly from visual
observations~\citep{zhai2024physical,shuai2025pugs,lin2025phys4dgen,dagli2025vomp}.
Most closely related to our setting, VoMP~\citep{dagli2025vomp} projects multi-view
visual features onto a volumetric grid and predicts Young's modulus, Poisson's
ratio, and density for individual voxels. Although this formulation substantially
reduces the need for manual annotation, it faces two fundamental limitations.

First, independent voxel-wise prediction overlooks the compositional structure of
real objects. Most objects contain only a small number of materials, such as a
wooden seat, metal legs, and rubber feet. Voxels made of the same material should
therefore exhibit similar mechanical properties and benefit from sharing evidence.
Independent regression does not explicitly impose this structure and may produce
noisy or inconsistent estimates across voxels belonging to the same material.
% Second, visual appearance does not uniquely determine material composition.
% Objects with similar geometry and appearance may be made from materials with
% substantially different mechanical properties. Figure~\ref{fig:motivation}
% illustrates this ambiguity using a plastic box and a metal container from the GVM dataset~\citep{dagli2025vomp}. Although the pair is nearly indistinguishable in the DINOv2 feature space,
% it is clearly separated in the RF representation: its pairwise-distance percentile
% increases from approximately \(0\%\) in the visual embedding to \(96\%\) in the RF
% embedding. This example highlights a fundamental limitation of vision-only
% estimation: similar visual representations may correspond to mechanically distinct
% materials.
Second, visual appearance does not uniquely determine material composition.
Objects with similar geometry and appearance may be made from materials with
substantially different mechanical properties. Figure~\ref{fig:motivation}
illustrates this ambiguity using a plastic box and a metal container from the GVM
dataset~\citep{dagli2025vomp}. Although the pair is nearly indistinguishable in
the DINOv2 feature space, it is clearly separated in the RF embedding: the
pairwise-distance percentile is approximately \(0\%\) in the visual embedding
but \(96\%\) in the RF embedding. This example highlights a fundamental limitation
of vision-only estimation: similar visual representations may correspond to
mechanically distinct materials.

To address these limitations, we introduce \textbf{ViWi}
(\textbf{Vi}sion Meets \textbf{Wi}Fi), an object-centric framework that combines
visual and radio-frequency information for volumetric mechanical-property
estimation. Rather than treating every voxel as an independent prediction problem,
ViWi represents an object using a compact set of latent material slots. Voxels are
softly assigned to these slots according to both visual compatibility and proximity
in a learned material-latent space. Each slot aggregates evidence from the voxels assigned to it and predicts a shared slot-level property prototype, from which voxel-level properties are reconstructed through soft assignments. This formulation promotes consistent predictions across voxels that share a material identity, including
spatially disconnected components.

ViWi further incorporates a compact RF descriptor obtained through physics-based
electromagnetic simulation. Because RF propagation depends on material-sensitive
properties such as permittivity and conductivity, it provides global composition
cues that may not be available from appearance alone. The RF descriptor conditions
the initial material slots, while the volumetric
visual features retain the spatial information required to localize materials.
Consequently, RF and visual evidence jointly influence material assignments and
property estimation while preserving their complementary roles: RF helps identify
which materials may be present, and vision determines where they occur.

Our contributions are summarized as follows:
\begin{itemize}
    \item We introduce ViWi, the first framework to combine visual observations
    with RF sensing for volumetric mechanical-property estimation, and construct
    physics-based RF descriptors for GVM through WiFi-band electromagnetic simulation.
    The complementary RF descriptor helps distinguish visually similar objects with
    different material compositions and physical properties.

    \item We reformulate volumetric property estimation as object-centric material
    decomposition. Our material-slot formulation groups voxels using visual and
    material-latent compatibility, allowing voxels with a shared material identity
    to aggregate evidence and receive coherent property predictions.

    \item We design an RF-conditioned material-slot initialization scheme based on
bounded, zero-initialized feature-wise linear modulation, enabling stable integration
of the global RF descriptor with voxel-level visual features for material grouping
and property estimation.

    \item ViWi improves over the prior state of the art on four of six
    volumetric mechanical-property metrics on GVM, while its vision-only variant
    improves all reported mass-estimation metrics on ABO-500. Controlled analyses
    further show that RF provides the largest gains when visual evidence is ambiguous.
\end{itemize}

\section{Related Work}
\label{sec:related_work}

\subsection{Mechanical-Property Estimation from Visual Observations}

Estimating mechanical properties from images or reconstructed 3D assets is
challenging because appearance provides only indirect evidence about physical
composition, while dense volumetric annotations are difficult to obtain. Early
approaches therefore relied on coarse material recognition or mappings from
semantic categories to physical parameters, which cannot recover spatially varying
Young's modulus, Poisson's ratio, and density.

Recent methods use pretrained representations and differentiable simulation to
infer richer physical attributes. NeRF2Physics~\citep{zhai2024physical} and
PUGS~\citep{shuai2025pugs} optimize NeRF or Gaussian-splat representations for
properties such as stiffness and density, but require per-object optimization and
provide limited access to object interiors. PhysDreamer~\citep{physdreamer},
DreamPhysics~\citep{dreamphysics}, Physics3D~\citep{physics3d}, and
PhysGaussian~\citep{physgaussian} combine generative priors or 3D representations
with physical simulation, but typically remain tied to a particular representation
or simulator.

Other approaches infer physical attributes from semantic object regions.
PhysGen~\citep{liu2024physgen}, PhysGen3D~\citep{chen2025physgen3d}, and
Phys4DGen~\citep{phys4dgen} use vision-language or segmentation-based cues to assign
materials and simulation parameters. Feed-forward methods such as
Pixie~\citep{pixie}, PhysSplat~\citep{zhao2025physsplat},
SOPHY~\citep{sophy}, and PhysX-3D~\citep{physx3d} improve inference efficiency,
but primarily estimate surface-oriented fields or generate physically augmented
assets rather than dense mechanical properties throughout an existing object's
volume.

% Most closely related to our setting, VoMP~\citep{dagli2025vomp} predicts volumetric
% Young's modulus, Poisson's ratio, and density from multi-view visual features and a
% learned material-latent space. Unlike its independent voxel-wise formulation, ViWi
% models the low-cardinality material composition of an object through material slots
% adapted from Slot Attention~\citep{slot_attention}, and incorporates complementary
% RF evidence to resolve visual ambiguity.

% Most closely related to our setting, VoMP~\citep{dagli2025vomp} predicts volumetric
% Young's modulus, Poisson's ratio, and density from multi-view visual features and a
% learned material-latent space, while the concurrent AdaVoMP~\citep{dagli2026adaptive}
% extends it with a sparse adaptive voxel structure that refines spatial resolution.
% Both estimate each voxel independently from appearance alone, so spatially disconnected
% parts that share a material, such as the four legs of a chair, are predicted separately,
% while visually indistinguishable materials remain ambiguous. ViWi instead models the
% low-cardinality material composition of an object through material slots adapted from
% Slot Attention~\citep{slot_attention}, allowing such parts to share material-slot
% representations and pool evidence across them, while incorporating complementary RF
% evidence to resolve visual ambiguity.

Most closely related, VoMP~\citep{dagli2025vomp} predicts volumetric Young's
modulus, Poisson's ratio, and density independently from multi-view visual features,
while AdaVoMP~\citep{dagli2026adaptive} improves spatial resolution using an adaptive
voxel structure but remains vision-only. ViWi instead models an object's
low-cardinality material composition using material slots adapted from Slot
Attention~\citep{slot_attention}, allowing spatially separated voxels that share a
material identity to aggregate evidence through a common slot. Complementary RF
information further provides material-sensitive cues that help resolve cases where
visual appearance alone is ambiguous.

\subsection{RF Sensing and Multimodal Material Perception}

RF sensing complements optical observations because electromagnetic propagation is
affected by material properties and remains informative under darkness, occlusion,
or privacy constraints. Prior work has applied RF signals to material
sensing~\citep{wifield}, activity recognition, and human pose
estimation~\citep{lee2022hupr,yu2023rfposeot,fan2024mmdiff}.
Vision--radar methods use cross-modal supervision or sensor fusion to improve pose
estimation under occlusion~\citep{lee2022hupr,knap2024radarcamera}, while RF-only
methods such as RFPose-OT~\citep{yu2023rfposeot} and
mmDiff~\citep{fan2024mmdiff} demonstrate that radio observations contain useful
structural information even when cameras are unreliable.

Existing vision--RF systems primarily target human sensing, pose estimation, or
material classification. In contrast, ViWi addresses dense volumetric estimation
of mechanical properties. It constructs a compact RF descriptor through
physics-based electromagnetic simulation and uses it to condition object-centric
material slots at both training and inference, allowing RF and visual evidence to
jointly influence material grouping and voxel-level property estimation.

\section{Method}
\label{sec:method}

\subsection{Problem Formulation and Overview}
\label{sec:method_overview}

Given multi-view observations of a 3D object, our goal is to estimate its volumetric
mechanical-property field. For each occupied voxel \(i\), the model predicts
\(\mathbf{y}_i=(E_i,\nu_i,\rho_i)\), where \(E_i\), \(\nu_i\), and \(\rho_i\)
denote Young's modulus, Poisson's ratio, and density, respectively.

Rather than predicting each voxel independently, ViWi represents an object using a
compact set of latent material slots. Each slot captures a shared material hypothesis,
aggregates evidence from the voxels assigned to it, and predicts a slot-level property
prototype. Voxel-level predictions are reconstructed through soft assignments to
these prototypes.

To reduce ambiguity from visual appearance, ViWi additionally uses an RF descriptor
obtained from physics-based electromagnetic simulation. The RF descriptor conditions
the material slots, allowing RF and visual evidence
to jointly influence material assignments and property estimation.

We build ViWi on the volumetric representation of VoMP~\citep{dagli2025vomp}. Its
pretrained geometry backbone aggregates multi-view DINOv2 features on a \(64^3\)
sparse voxel grid and produces \(\mathbf{x}_i\in\mathbb{R}^{768}\) for each occupied
voxel. We keep the geometry backbone and MatVAE frozen. MatVAE maps mechanically valid
property triplets to a two-dimensional material-latent space and decodes them back to
\((E,\nu,\rho)\).

\begin{figure*}[t]
    \centering
    \includegraphics[width=\textwidth]{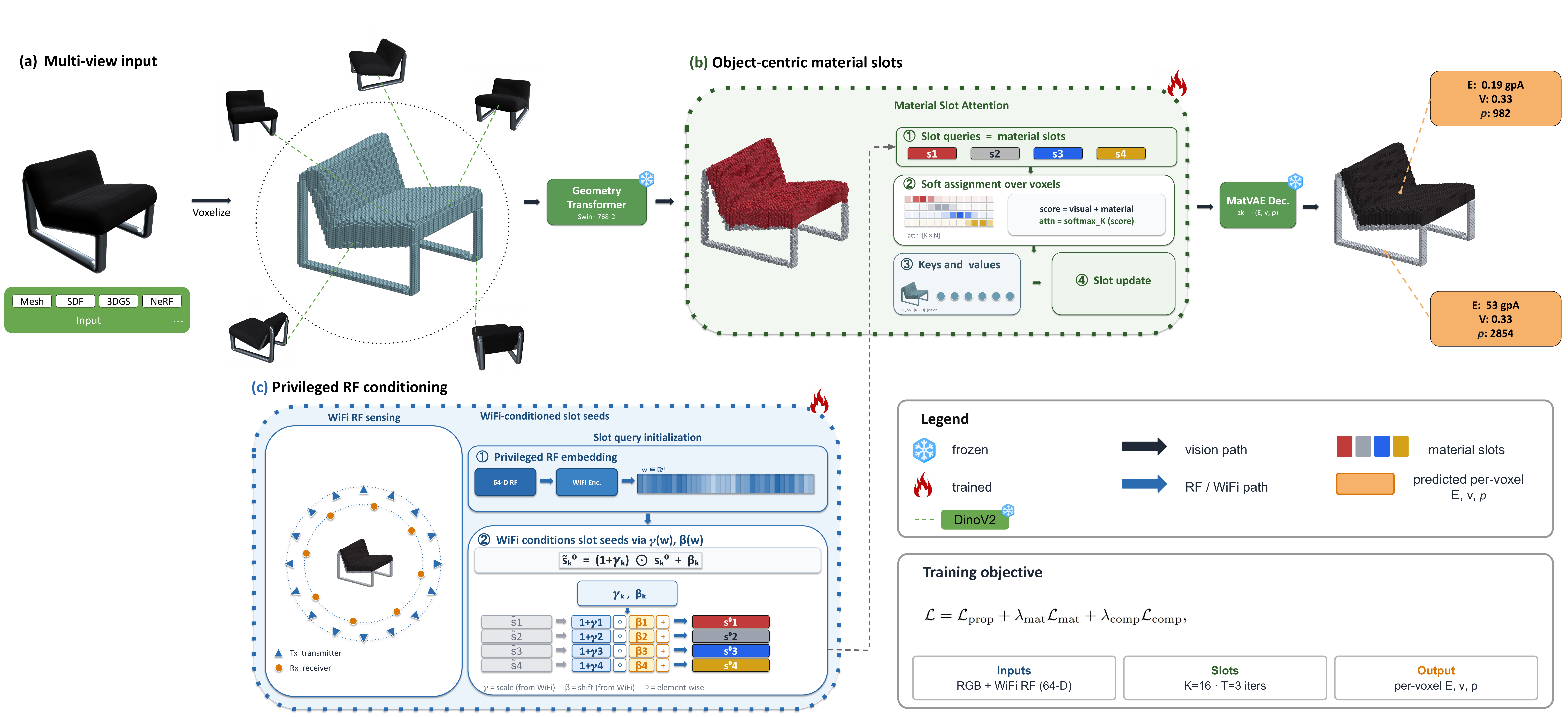}
\caption{
\textbf{Overview of ViWi.}
Multi-view RGB features are aggregated on a sparse voxel grid and grouped into
material slots using visual and material-latent compatibility. Each slot material
latent is decoded by the frozen MatVAE into \(E\), \(\nu\), and \(\rho\), and
voxel-level properties are reconstructed through soft assignments. A simulated
\(64\)-D RF descriptor conditions the initial material-slot states through bounded
feature-wise modulation, providing complementary object-level material information.
}
    \label{fig:method_overview}
    % \hspace{10 pt}
\end{figure*}

\subsection{Object-Centric Material Slot Attention}

% \paragraph{Material-slot representation.}

% Building on Slot Attention~\citep{slot_attention}, ViWi represents each object
% using \(K\) latent material slots. Slot \(k\) consists of a feature embedding
% \(\mathbf{s}_k\in\mathbb{R}^{d_s}\) and a two-dimensional slot material latent
% \(\mathbf{m}_k\in\mathbb{R}^{2}\). The feature embedding stores evidence aggregated
% from voxel features, whereas the slot material latent represents the slot's current
% hypothesis about the corresponding mechanical properties.
% The two-dimensional material latent provides each slot with an explicit
% material hypothesis that guides voxel-to-slot assignment through compatibility
% in the pretrained material-latent space.

\paragraph{Material-slot representation.}
Building on Slot Attention~\citep{slot_attention}, ViWi represents each object using
\(K\) latent material slots. Each slot contains a feature embedding
\(\mathbf{s}_k\in\mathbb{R}^{d_s}\) that aggregates voxel-level visual evidence and
a two-dimensional material-latent seed
\(\bar{\mathbf{m}}_k\in\mathbb{R}^{2}\) in the pretrained MatVAE latent space.
Together, these two representations allow each slot to capture both visual evidence
and a hypothesis about material identity. 

% The material-latent seed provides the material-compatibility cue for voxel-to-slot
% assignment, while the slot feature embedding \(\mathbf{s}_k\) and voxel feature
% embedding provide the visual-compatibility cue.

\paragraph{Voxel and material representations.}

The frozen geometry backbone produces voxel features
\[
\mathbf{X}=[\mathbf{x}_1,\ldots,\mathbf{x}_N]^\top,
\]
where \(\mathbf{x}_i\) represents the visual information associated with occupied
voxel \(i\). Learned projections transform these features into voxel keys
\(\mathbf{k}_i\) and values \(\mathbf{v}_i\), which are used for voxel-to-slot
assignment and evidence aggregation, respectively. In parallel, a separate prediction
head maps each voxel feature to a vision-based voxel material latent
\(\widehat{\mathbf{m}}_i\in\mathbb{R}^{2}\) in the pretrained MatVAE latent space.
This latent provides a material-compatibility cue for voxel-to-slot assignment rather
than serving as the final voxel-level property prediction.

\paragraph{Material-aware assignment.}

At each refinement iteration, ViWi computes a soft voxel-to-slot assignment.
For voxel \(i\) and slot \(k\), the assignment combines visual compatibility with
proximity in the material-latent space:

\begin{equation}
\label{eq:assignment_score}
\begin{aligned}
a^{\mathrm{vis}}_{ik}
    &= \frac{q(\mathbf{s}_k)^\top \mathbf{k}_i}{\sqrt{d_a}},\\
a^{\mathrm{mat}}_{ik}
    &= -\left\|\mathbf{m}_k-\widehat{\mathbf{m}}_i\right\|_2^2,\\
a_{ik}
    &= \lambda_{\mathrm{vis}}a^{\mathrm{vis}}_{ik}
     + \lambda_{\mathrm{mat}}a^{\mathrm{mat}}_{ik},
\end{aligned}
\end{equation}

where \(q(\cdot)\) is the slot-query projection, \(d_a\) is the attention
dimension, and \(\lambda_{\mathrm{vis}}\) and \(\lambda_{\mathrm{mat}}\) are
learned non-negative weights. We normalize the scores across slots:

\begin{equation}
\label{eq:slot_assignment}
\alpha_{ik}
=
\frac{\exp(a_{ik})}
{\sum_{k'=1}^{K}\exp(a_{ik'})},
\qquad
\sum_{k=1}^{K}\alpha_{ik}=1.
\end{equation}

The assignment weight \(\alpha_{ik}\) represents the degree to which voxel \(i\)
is assigned to slot \(k\). Normalization across slots creates competition among
material hypotheses and encourages the slots to specialize in distinct material
identities.

\paragraph{Iterative evidence aggregation.}

Each slot aggregates the value features of the voxels assigned to it using

\begin{equation}
\label{eq:slot_update}
\mathbf{u}_k
=
\sum_{i=1}^{N}\alpha_{ik}\mathbf{v}_i .
\end{equation}

The aggregated feature \(\mathbf{u}_k\) updates the slot feature embedding
\(\mathbf{s}_k\) through a gated recurrent unit followed by a residual
feed-forward network. Repeating this process allows the updated slot feature
embeddings to refine the subsequent voxel-to-slot assignments, which in turn
determine the voxel evidence aggregated by each slot. After the final refinement
iteration, the slot material latent is obtained from the updated slot feature
embedding through a residual prediction added to its initial material-latent seed.

\paragraph{Slot-level decoding and voxel reconstruction.}

Each final slot material latent
\(\mathbf{m}_k\in\mathbb{R}^{2}\) is mapped by the frozen MatVAE decoder
\(g_{\mathrm{mat}}:\mathbb{R}^{2}\rightarrow\mathbb{R}^{3}\) to a
slot-level property prototype,

\[
\mathbf{p}_k
=
g_{\mathrm{mat}}(\mathbf{m}_k)
=
(E_k,\nu_k,\rho_k)
\in\mathbb{R}^{3}.
\]

Collecting all slot predictions gives
\(\mathbf{P}=[\mathbf{p}_1,\ldots,\mathbf{p}_K]^\top\in\mathbb{R}^{K\times 3}\).
The final voxel-level prediction is reconstructed using the soft voxel-to-slot
assignment:

\begin{equation}
\label{eq:voxel_reconstruction}
\widehat{\mathbf{y}}_i
=
\sum_{k=1}^{K}\alpha_{ik}\mathbf{p}_k
=
\boldsymbol{\alpha}_i^\top\mathbf{P},
\qquad
\boldsymbol{\alpha}_i=
[\alpha_{i1},\ldots,\alpha_{iK}]^\top.
\end{equation}

This reconstruction promotes consistency among voxels assigned to the same
material slot while allowing smooth transitions between material slots through soft assignments.

% \subsection{Physics-Based RF Representation}
% \label{sec:rf_representation}

% To complement visual appearance with material-sensitive information, ViWi constructs
% a compact RF descriptor for each GVM object through physics-based electromagnetic
% simulation. Each mechanical material category is mapped to representative
% electromagnetic parameters, namely relative permittivity \(\epsilon_r\) and
% conductivity \(\sigma\), which determine how RF waves interact with the object.

% The simulated propagation paths are aggregated into a fixed-size global descriptor.
% For each transmitter, we compute four statistics over all receivers: total scattered
% power, the fraction of receivers reached by a valid path, maximum single-path gain,
% and RMS delay spread. Stacking these statistics across \(T\) transmitters gives

% \begin{equation}
% \label{eq:rf_descriptor}
% \mathbf{r}\in\mathbb{R}^{D},
% \qquad
% D=T\times F,
% \end{equation}

% where \(T\) is the number of transmitters and \(F\) is the number of
% per-transmitter RF statistics. The resulting descriptor
% captures a global, material-sensitive fingerprint of the object but does not encode
% the spatial locations of individual materials. We therefore use it to condition the
% initial material slots, while voxel-level visual features provide spatial
% localization. Detailed electromagnetic simulation settings are provided in the
% supplementary material.

\subsection{Physics-Based RF Representation}
\label{sec:rf_representation}

To complement visual appearance with material-sensitive information, ViWi constructs
a compact RF descriptor for each GVM object through physics-based electromagnetic
simulation. Each mechanical material category is mapped to representative
electromagnetic parameters, namely relative permittivity \(\epsilon_r\) and
conductivity \(\sigma\), which determine how RF waves interact with the object.

The simulated propagation paths are aggregated into a fixed-size global descriptor.
For each transmitter, we compute four statistics over all receivers: total scattered
power, the fraction of receivers reached by a valid path, maximum single-path gain,
and RMS delay spread. Stacking these statistics across \(N_{\mathrm{tx}}\)
transmitters gives

\begin{equation}
\label{eq:rf_descriptor}
\mathbf{r}\in\mathbb{R}^{D},
\qquad
D=N_{\mathrm{tx}}\times F,
\end{equation}

where \(N_{\mathrm{tx}}\) is the number of transmitters and \(F\) is the number of
per-transmitter RF statistics. The resulting RF descriptor captures global,
material-sensitive information about the object but does not encode the spatial
locations of individual materials. We therefore use it to condition the initial
material-slot states, while voxel-level visual features provide spatial localization.
Detailed electromagnetic simulation settings are provided in the supplementary
material.

\subsection{RF-Conditioned Material Slots}
\label{sec:rf_conditioning}

The RF descriptor provides global evidence about material composition but does not
encode the spatial locations of individual materials. We therefore use it to
condition the initial material-slot states, while voxel-level visual features
provide the spatial information required for voxel-to-slot assignment.

The standardized RF descriptor \(\mathbf{r}\) is encoded by a multilayer perceptron
into an RF embedding
\(\mathbf{z}_{\mathrm{RF}}\in\mathbb{R}^{d_{\mathrm{RF}}}\). From this embedding,
the network predicts feature-wise scale and shift parameters for both the slot
feature embedding and the material-latent seed. Following feature-wise linear
modulation (FiLM)~\citep{perez2018film}, the initial state of slot \(k\) is
conditioned as

\begin{equation}
\label{eq:rf_conditioning}
\begin{aligned}
\widetilde{\mathbf{s}}_k^{\,0}
    &=
    \left(1+\boldsymbol{\gamma}_k^{s}\right)
    \odot \mathbf{s}_k^{0}
    +\boldsymbol{\beta}_k^{s},
&\qquad
\widetilde{\mathbf{m}}_k^{\,0}
    &=
    \left(1+\boldsymbol{\gamma}_k^{m}\right)
    \odot \mathbf{m}_k^{0}
    +\boldsymbol{\beta}_k^{m},
\end{aligned}
\end{equation}

where the superscript \(0\) denotes the slot state before the first refinement
iteration. Each modulation parameter is obtained from a linear projection of
\(\mathbf{z}_{\mathrm{RF}}\) followed by \(c\tanh(\cdot)\), which bounds its
magnitude by \(c\). This prevents the global RF descriptor from overwhelming the
spatially resolved visual features. Modulating the material-latent seed
\(\mathbf{m}_k^{0}\) directly affects the material-compatibility score
\(a^{\mathrm{mat}}_{ik}\) in Eq.~\eqref{eq:assignment_score}, while modulating
the slot feature embedding \(\mathbf{s}_k^{0}\) affects the visual-compatibility
score.

Because the effect of the initial modulation may be attenuated during iterative
slot refinement, we additionally apply a bounded RF-dependent correction after the
final iteration:

\begin{equation}
\Delta\mathbf{m}_k^{\mathrm{RF}}
=
c\tanh\!\left(\mathbf{W}_{r}\mathbf{z}_{\mathrm{RF}}\right).
\end{equation}

This correction is added to the final slot material latent before decoding,
providing a direct RF-dependent path to the slot-level property prototype. RF therefore
acts only on slot-level quantities, while voxel-level visual features continue to
provide the spatial evidence required to localize materials.

All RF-conditioning projections are initialized with zero weights and biases.
Thus, at the beginning of multimodal training,
\(\boldsymbol{\gamma}=\boldsymbol{\beta}=\mathbf{0}\) and
\(\Delta\mathbf{m}_k^{\mathrm{RF}}=\mathbf{0}\), so the RF pathway initially
reduces to the identity transformation. The remaining parameters are initialized
from the trained vision-only variant, ViWi\(^{\dagger}\), allowing RF conditioning
to be learned gradually from the vision-only solution.

\subsection{Learning Objectives}
\label{sec:learning_objectives}

Let \(\mathbf{y}_i=(E_i,\nu_i,\rho_i)\) and
\(\widehat{\mathbf{y}}_i\) denote the ground-truth and predicted properties of
occupied voxel \(i\). We use an \(\ell_1\) property loss. We further align the
vision-based voxel material latent \(\widehat{\mathbf{m}}_i\) with a detached
target \(\mathbf{m}^{\mathrm{enc}}_i\) from the frozen MatVAE encoder, stabilizing
the material-compatibility cue used for voxel-to-slot assignment. In parallel,
the RF embedding is supervised with the object-level material composition,
encouraging it to encode which materials are present.

For object \(b\) in batch \(\mathcal{B}\), let
\(\mathbf{h}_b\in\mathbb{R}^{C}\) denote its ground-truth material-category
distribution over \(C\) categories and
\(\boldsymbol{\ell}_b\in\mathbb{R}^{C}\) the predicted composition logits.
The losses are

\begin{equation}
\begin{aligned}
\mathcal{L}_{\mathrm{prop}}
&=\frac{1}{N}\sum_{i=1}^{N}
\left\|\widehat{\mathbf{y}}_i-\mathbf{y}_i\right\|_1,
\qquad
\mathcal{L}_{\mathrm{mat}}
=\frac{1}{N}\sum_{i=1}^{N}
\left\|\widehat{\mathbf{m}}_i-\mathbf{m}^{\mathrm{enc}}_i\right\|_1,\\
\mathcal{L}_{\mathrm{comp}}
&=-\frac{1}{|\mathcal{B}|}
\sum_{b\in\mathcal{B}}\sum_{c=1}^{C}
h_{b,c}\log\operatorname{softmax}(\boldsymbol{\ell}_b)_c,
\qquad
\mathcal{L}
=\mathcal{L}_{\mathrm{prop}}
+\lambda_{\mathrm{mat}}\mathcal{L}_{\mathrm{mat}}
+\lambda_{\mathrm{comp}}\mathcal{L}_{\mathrm{comp}}.
\end{aligned}
\end{equation}

\section{Experiments}
\label{sec:experiments}

We evaluate ViWi on volumetric mechanical-property estimation and assess whether
the predicted density fields also improve downstream object-mass estimation.
Experiments are conducted on GVM~\citep{dagli2025vomp} for volumetric estimation
of Young's modulus, Poisson's ratio, and density, and on
ABO-500~\citep{collins2022abo} for mass estimation derived from the predicted
volumetric density field. Complete dataset descriptions, including the construction of RF descriptors for GVM,
together with evaluation metrics, implementation details, and training settings are
provided in the supplementary material.

\begin{table}[t]
\centering
% \caption{
% Volumetric mechanical-property estimation on the GVM hold-out test set of
% 166 objects. Following VoMP~\citep{dagli2025vomp}, we report Average Log
% Displacement Error (ALDE) and Average Log Relative Error (ALRE) for Young's
% modulus, and Average Displacement Error (ADE) and Average Relative Error (ARE)
% for Poisson's ratio and density. VoMP results are reproduced using the authors'
% original pretrained weights to ensure a fair comparison. Lower is better. Best results are shown in
% \textbf{bold}, and second-best results are \underline{underlined}.
% }
% \caption{
% Volumetric mechanical-property estimation on the GVM hold-out test set.
% We report ALDE and ALRE for Young's modulus, and ADE and ARE for Poisson's ratio
% and density, following VoMP~\citep{dagli2025vomp}. VoMP results are reproduced
% using the authors' pretrained weights to ensure a fair comparison. Lower is better; best and second-best results
% are shown in \textbf{bold} and \underline{underlined}, respectively.
% }
\caption{
Volumetric mechanical-property estimation on the GVM hold-out test set.
We report ALDE and ALRE for Young's modulus, and ADE and ARE for Poisson's ratio
and density, following VoMP~\citep{dagli2025vomp}. Errors are computed per object
and then averaged across the test set. VoMP results are reproduced using the
authors' pretrained weights to ensure a fair comparison. Lower is better; best results are
shown in \textbf{bold}.
}
\label{tab:gvm_comparison}
\vspace{4pt}
\footnotesize
\resizebox{\textwidth}{!}{%
\begin{tabular}{@{}l *{6}{c} @{}}
\toprule
% \rowcolor{lightgreen}
\multirow{2}{*}{Method}
& \multicolumn{2}{c}{Young's Modulus (\(E\)) [Pa]}
& \multicolumn{2}{c}{Poisson's Ratio (\(\nu\))}
& \multicolumn{2}{c}{Density (\(\rho\)) [kg/m\(^3\)]} \\
\cmidrule(lr){2-3}
\cmidrule(lr){4-5}
\cmidrule(lr){6-7}
% \rowcolor{lightgreen}
& ALDE (\(\downarrow\))
& ALRE (\(\downarrow\))
& ADE (\(\downarrow\))
& ARE (\(\downarrow\))
& ADE (\(\downarrow\))
& ARE (\(\downarrow\)) \\
\midrule
NeRF2Physics - CVPR 2024~\citep{zhai2024physical}
& 2.8000 {\scriptsize (\(\pm\)1.05)}
& 0.1346 {\scriptsize (\(\pm\)0.05)}
& --
& --
& 1432.03 {\scriptsize (\(\pm\)964.88)}
& 1.0365 {\scriptsize (\(\pm\)0.63)} \\

PUGS - ICRA 2025~\citep{shuai2025pugs}
& 3.3942 {\scriptsize (\(\pm\)1.72)}
& 0.1688 {\scriptsize (\(\pm\)0.10)}
& --
& --
& 3568.22 {\scriptsize (\(\pm\)2839.13)}
& 3.2429 {\scriptsize (\(\pm\)3.56)} \\

Phys4DGen* - ACM MM 2025~\citep{phys4dgen}
& 4.8967 {\scriptsize (\(\pm\)3.17)}
& 0.2227 {\scriptsize (\(\pm\)0.14)}
& 0.0407 {\scriptsize (\(\pm\)0.04)}
& 0.1467 {\scriptsize (\(\pm\)0.18)}
& 1865.57 {\scriptsize (\(\pm\)2176.90)}
& 1.4394 {\scriptsize (\(\pm\)2.35)} \\

% VoMP~\citep{dagli2025vomp}
% & \underline{0.3925}
% & \underline{0.0441}
% & \underline{0.0254}
% & \underline{0.0859}
% & \textbf{119.61}
% & \underline{0.0881} \\

% \midrule
% \textbf{ViWi (Ours)}
% & \textbf{0.1667}
% & \textbf{0.0188}
% & \textbf{0.0143}
% & \textbf{0.0487}
% & \underline{137.80}
% & \textbf{0.0821} \\
% \bottomrule
% \end{tabular}%
% }

VoMP - ICLR 2026~\citep{dagli2025vomp}
& {0.3952 {\scriptsize (\(\pm\)0.30)}}
& {0.0427 {\scriptsize (\(\pm\)0.038)}}
& {0.0245 {\scriptsize (\(\pm\)0.007)}}
& {0.0842 {\scriptsize (\(\pm\)0.029)}}
& {153.06 {\scriptsize (\(\pm\)174)}}
& {0.0898 {\scriptsize (\(\pm\)0.071)}} \\

AdaVoMP - ICML 2026~\citep{dagli2026adaptive}
& {0.3278 {\scriptsize (\(\pm\)0.26)}}
& {0.0340 {\scriptsize (\(\pm\)0.03)}}
& {0.0205 {\scriptsize (\(\pm\)0.01)}}
& {0.0680 {\scriptsize (\(\pm\)0.03)}}
& \textbf{127.31 {\scriptsize (\(\pm\)150.83)}}
& \textbf{0.0842 {\scriptsize (\(\pm\)0.07)}} \\

\midrule
\textbf{ViWi (Ours)}
& \textbf{0.1797 {\scriptsize (\(\pm\)0.21)}}
& \textbf{0.0196 {\scriptsize (\(\pm\)0.026)}}
& \textbf{0.0151 {\scriptsize (\(\pm\)0.015)}}
& \textbf{0.0515 {\scriptsize (\(\pm\)0.052)}}
& {198.22 {\scriptsize (\(\pm\)340.85)}}
& {0.1034 {\scriptsize (\(\pm\)0.11)}} \\
\bottomrule
\end{tabular}%
}

\vspace{2pt}
\end{table}

\begin{wraptable}{r}{0.47\columnwidth}
\vspace{-35pt}
\centering
\caption{
{Mass estimation on ABO-500.}
Results on 100 real test objects. The upper block reports published results,
while the lower block reports results reproduced under our fixed evaluation
protocol.
}
\label{tab:abo500_mass}
\vspace{-2pt}

\resizebox{\linewidth}{!}{%
\begin{tabular}{@{}lcccc@{}}
\toprule
\textbf{Method}
& \textbf{ALDE} \(\downarrow\)
& \textbf{ADE [kg]} \(\downarrow\)
& \textbf{ARE} \(\downarrow\)
& \textbf{MnRE} \(\uparrow\) \\
\midrule

\multicolumn{5}{@{}l}{\textit{Reported in original papers}} \\
\addlinespace[2pt]

NeRF2Physics~\cite{nerf2physics}
& 0.736 & 12.725 & 1.040 & 0.564 \\

PUGS~\cite{pugs}
& 0.661 & 9.461 & 0.767 & 0.576 \\

Phys4DGen$^{\star}$~\cite{phys4dgen}
& 0.664 & 9.961 & 0.825 & 0.566 \\

VoMP~\cite{vomp}
& 0.631 & 8.433 & 0.887 & 0.576 \\

\midrule

\multicolumn{5}{@{}l}{\textit{Reproduced under our fixed protocol}} \\
\addlinespace[2pt]

VoMP~\cite{vomp}
& 1.170 & 37.372 & 3.121 & 0.395 \\

\textbf{ViWi}$^{\dagger}$ (Ours)
& \textbf{1.113}
& \textbf{20.181}
& \textbf{2.538}
& \textbf{0.415} \\

\bottomrule
\end{tabular}%
}
\vspace{-8pt}
\end{wraptable}

\subsection{Main Results}
\label{sec:main_results}

Table~\ref{tab:gvm_comparison} compares ViWi with prior methods on the GVM
benchmark, all evaluated on the same \(64^3\) grid; for the concurrent
AdaVoMP~\citep{dagli2026adaptive} we quote its own reported \(64^3\) operating
point, obtained by halting its coarse-to-fine decoder early. ViWi is best on
four of the six metrics. Against VoMP~\citep{dagli2025vomp}, ViWi reduces
Young's modulus ALDE/ALRE from \(0.3952/0.0427\) to \(0.1797/0.0196\), i.e.\ by
\(54.5\%\) and \(54.1\%\); this matters because \(E\) spans several orders of
magnitude and is therefore hard to infer from visual appearance alone. For
Poisson's ratio, ADE/ARE fall from \(0.0245/0.0842\) to \(0.0151/0.0515\)
(\(38.4\%\) and \(38.8\%\)).

ViWi also outperforms AdaVoMP on \(E\) (\(45.2\%\) lower ALDE, \(42.4\%\) lower
ALRE) and \(\nu\) (\(26.3\%\) lower ADE, \(24.3\%\) lower ARE). Both numbers are
read at \(64^3\), yet AdaVoMP enters the comparison with a considerably stronger
recipe: a \(0.6\)B-parameter adaptive-voxel model using DINOv3 ViT-H\(+\)/16
features and depth-attenuated multi-view averaging, evaluated at \(64^3\) by
halting its coarse-to-fine decoder early. ViWi instead predicts natively on
VoMP's fixed \(64^3\) grid from DINOv2 features with uniform view averaging. Our
gains thus stem from material grouping and RF conditioning rather than a
stronger backbone or finer discretisation, and the two directions are
complementary.

Density is the only property for which ViWi has a higher mean error, with
ADE/ARE of \(198.22/0.1034\) compared with \(153.06/0.0898\) for VoMP.
The gap is driven by a small number of outlier objects rather than a systematic
performance loss: the two models are effectively tied on the median test object
(ADE \(113.1\) vs.\ \(114.0\)), the ten worst objects account for \(93\%\) of
the mean gap, and a paired Wilcoxon signed-rank test on the per-object errors of
the 148 test objects finds no significant difference (\(p=0.24\)).
Density is also the only property whose metric can expose such a tail: Young's
modulus is scored in log space, which compresses even order-of-magnitude
discrepancies, and Poisson's ratio is bounded to \([0,0.49]\) in GVM, whereas
density is scored linearly and spans \(102\)--\(19{,}800\)~kg/m\(^3\).

\begin{table}[b]
\centering
\caption{
\textbf{Material-slot attention ablation on GVM.}
Comparison of VoMP, ViWi\(^{\dagger}\) (Vision Only), and the full ViWi model.
Lower is better for all metrics.
}
\label{tab:material_slot_ablation}
\vspace{2pt}

\scriptsize
\setlength{\tabcolsep}{2.5pt}
\resizebox{0.8\linewidth}{!}{%
\begin{tabular}{@{}lcccccc@{}}
\toprule
Method
& \multicolumn{2}{c}{Young's Modulus (\(E\))}
& \multicolumn{2}{c}{Poisson's Ratio (\(\nu\))}
& \multicolumn{2}{c}{Density (\(\rho\))} \\
\cmidrule(lr){2-3}
\cmidrule(lr){4-5}
\cmidrule(lr){6-7}
&
ALDE
& ALRE
& ADE
& ARE
& ADE
& ARE \\
\midrule

VoMP~\citep{dagli2025vomp}
& 0.3952 {\tiny (\(\pm\)0.3012)}
& 0.0427 {\tiny (\(\pm\)0.0381)}
& 0.0245 {\tiny (\(\pm\)0.0072)}
& 0.0842 {\tiny (\(\pm\)0.0288)}
& \textbf{153.06 {\tiny (\(\pm\)173.7823)}}
& \textbf{0.0898 {\tiny (\(\pm\)0.0705)}} \\

ViWi\(^{\dagger}\)
& \textbf{0.1853 {\tiny (\(\pm\)0.2026)}}
& \textbf{0.0202 {\tiny (\(\pm\)0.0249)}}
& \textbf{0.0146 {\tiny (\(\pm\)0.0151)}}
& \textbf{0.0503 {\tiny (\(\pm\)0.0558)}}
& {238.71 {\tiny (\(\pm\)304.1573)}}
& {0.1243 {\tiny (\(\pm\)0.1198)}} \\

% ViWi
% & \textbf{0.1797 {\tiny (\(\pm\)0.2121)}}
% & \textbf{0.0196 {\tiny (\(\pm\)0.0257)}}
% & \textbf{0.0151 {\tiny (\(\pm\)0.0152)}}
% & \textbf{0.0515 {\tiny (\(\pm\)0.0516)}}
% & \textbf{198.22 {\tiny (\(\pm\)340.8454)}}
% & \textbf{0.1034 {\tiny (\(\pm\)0.1080)}} \\

\bottomrule
\end{tabular}}
\end{table}

We further evaluate whether the predicted density fields yield accurate mass
estimates on the real-object ABO-500 benchmark. ViWi\(^{\dagger}\) denotes our
vision-only variant: RF is unavailable here, since our descriptors are simulated
from per-voxel electromagnetic properties while ABO-500 provides only a single
ground-truth mass. Converting density to mass also requires a volume estimate
that prior work leaves unspecified, so published numbers are not directly
comparable. We therefore re-evaluate VoMP and ViWi\(^{\dagger}\) under one
identical protocol. No protocol of the form \(m=c\,\rho V\) recovers VoMP's
reported ALDE of \(0.631\) from its released weights (best attainable
\(1.094\)), so the gap lies in unpublished evaluation details rather than in the
weights. Under our protocol (Table~\ref{tab:abo500_mass}), ViWi\(^{\dagger}\)
improves on all four metrics, reducing ADE by \(46.0\%\).

% \begin{figure}[t]
%     \centering
%     \includegraphics[width=\textwidth]{figures/abo_mass_examples.pdf}
%     \caption{
%     \textbf{Qualitative mass estimation on ABO-500.}
%     Representative test objects with ground-truth mass and estimates from
%     VoMP~\citep{dagli2025vomp} and ViWi\(^{\dagger}\).
%     }
%     \label{fig:abo_mass_examples}
% \end{figure}

\begin{figure}[t]
  \centering
  \includegraphics[width=\linewidth]{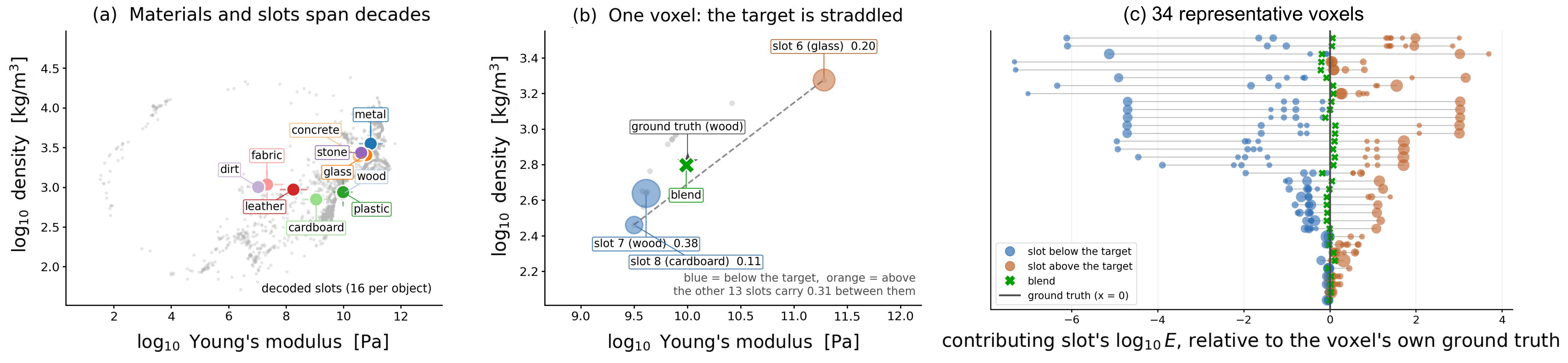}
\caption{
{Why soft voxel-to-slot assignment helps.}
\textbf{(a)} Decoded material slots span the broad range of mechanical properties.
\textbf{(b)} A voxel can combine softer and stiffer slots to match a target not
represented by any single slot; circle size is the attention weight.
\textbf{(c)} This bracketing behavior occurs systematically across voxels, shown for
\(34\) representative ones.
\textbf{(d)} Replacing soft assignment with straight-through Gumbel one-hot assignment
increases overall relative error from \(6.42\%\) to \(7.72\%\).
}
  \label{fig:bracketing}
\end{figure}

\subsection{Ablation Studies}

\paragraph{Where does RF help? Gain versus visual difficulty.}
\label{sec:ablation_difficulty}

\begin{wrapfigure}{r}{0.46\textwidth}
    \vspace{-18pt}
    \centering
    \includegraphics[width=\linewidth]{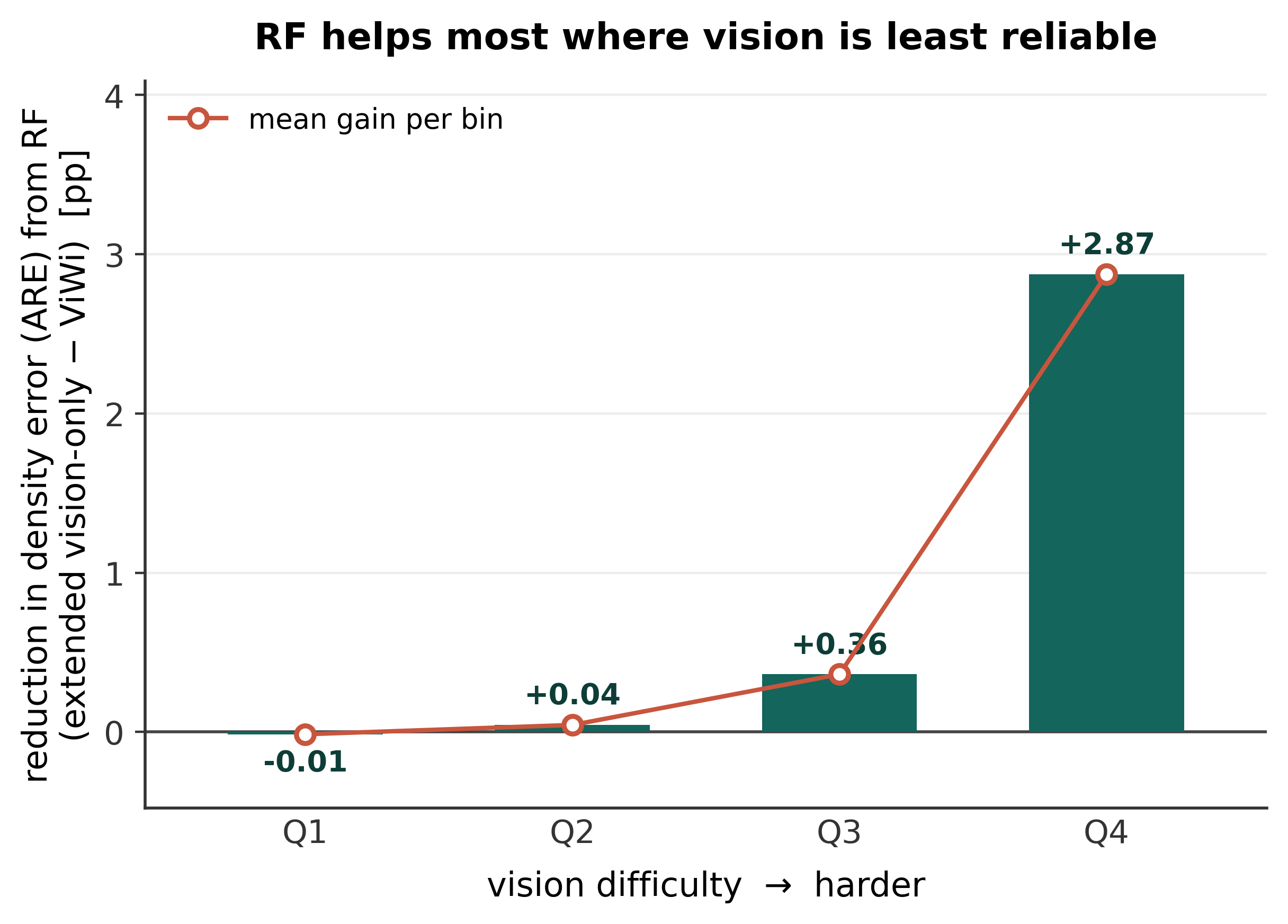}
\caption{RF gain versus visual difficulty. Objects are ranked by the overall error of
ViWi\(^{\ddagger}\) (ViWi\(^{\dagger}\) trained to ViWi's step budget with RF withheld)
and split into four equal-count quartiles. Bars give the mean reduction in density
error (ARE) in percentage points; positive is better.}

    \label{fig:wifi_gain_difficulty}
    \vspace{-39pt}
\end{wrapfigure}

We rank the \(305\) validation and test objects by the overall error of
ViWi\(^{\ddagger}\) and split them into four equal-count quartiles (\(Q_1\) easiest to
\(Q_4\) hardest). ViWi\(^{\ddagger}\) is ViWi\(^{\dagger}\) trained for the same number
of steps as ViWi with the RF input withheld, so the gap between them reflects RF rather
than extra optimisation. Figure~\ref{fig:wifi_gain_difficulty} reports the resulting
reduction in density error: RF is neutral on the easiest quartile (\(-0.01\) pp) and
reaches \(+2.87\) pp on the hardest, where \(63\%\) of objects improve. RF thus adds complementary material information for density where visual appearance is least reliable, while having little effect where vision already suffices.

\paragraph{Why soft voxel-to-slot assignment helps.}
ViWi predicts each voxel as a weighted mixture of slots rather than assigning it to
a single slot. The slots cover the wide range of materials in GVM
(Fig.~\ref{fig:bracketing}(a)), but a single slot rarely matches a given voxel
exactly. Mixing a softer and a stiffer slot allows the prediction to fall between
them and land closer to the ground truth than either slot alone
(Fig.~\ref{fig:bracketing}(b)). This is the common case rather than a special one
(Fig.~\ref{fig:bracketing}(c)): for \(74.1\%\) of voxels, the true values lie
between the contributing slots for all three properties simultaneously; for
\(83.6\%\), the mixture outperforms its best single slot; and the error for these
voxels is lower (\(0.041\) versus \(0.072\)). Forcing a hard choice with a
straight-through Gumbel variant, all else being equal, raises the overall relative
error---the mean of the relative errors on \(\log_{10}E\), \(\nu\), and \(\rho\)---
from \(6.42\%\) to \(7.72\%\) at matched sample counts.

\paragraph{Contribution of material slot attention.}
\label{sec:ablation_material_slots}

To isolate the material-slot formulation, we compare VoMP's voxel-wise regression
with ViWi\(^{\dagger}\) (Vision Only), which shares the frozen geometry backbone
and MatVAE decoder but receives no RF input. As shown in
Table~\ref{tab:material_slot_ablation}, material-slot attention substantially
improves both metrics for Young's modulus and Poisson's ratio, by approximately
\(53\%\) and \(40\%\), respectively. Density is the exception, although its gap is
largely driven by a few outlier objects as discussed in Sec.~\ref{sec:main_results};
the full ViWi model recovers much of this loss. These results
support grouping voxels through shared material slots rather than predicting each
voxel independently.

% \begin{figure}[t]
% \centering
% \includegraphics[width=\linewidth]{figures/wifi_gain_difficulty.png}
% \caption{
% \textbf{RF gain concentrates on visually hard objects.} Objects
% (\(305\), val+test) are binned into equal-count quintiles by ViWi (Vision Only)
% error (\(Q_1\) easiest \(\rightarrow\) \(Q_5\) hardest); each bar is the mean RF
% gain (ViWi (Vision Only) error \(-\) ViWi error; positive \(=\) RF helps), with
% error bars over \(\sim\!61\) objects per bin. Below each bar we annotate the number
% of objects in the bin (\(n\)) and the percentage of those objects for which RF
% lowered the error (``\% helped''). \textbf{Left:} overall error.
% \textbf{Right:} density, the most material-driven property. The gain increases with
% visual difficulty and is largest on the hardest quintile (\(+1.63\)~pp overall,
% \(+4.78\)~pp density). On the easiest density bins the gain is slightly negative,
% indicating that when vision already suffices the RF cue is unnecessary; RF becomes
% decisive as ambiguity grows.
% }
% \label{fig:wifi_gain_difficulty}
% \end{figure}

\begin{figure*}[t]
    \centering
    \includegraphics[width=\textwidth]{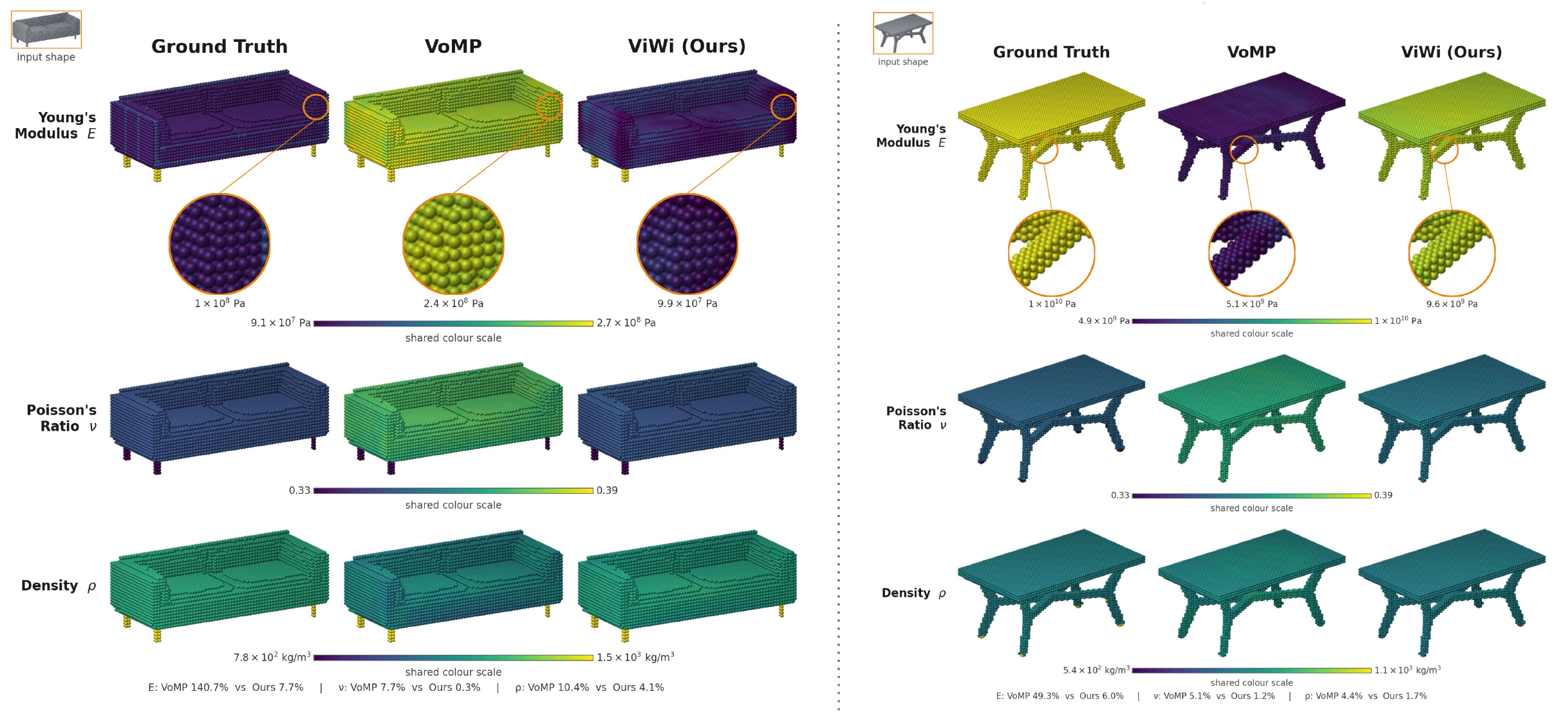}
\caption{
{Qualitative comparison of volumetric mechanical-property estimation.}
Ground truth, VoMP~\citep{dagli2025vomp}, and ViWi predictions for \(E\), \(\nu\), and
\(\rho\) on two representative GVM test objects. Percentages are per-object mean
relative errors, with \(E\) measured in linear \(\mathrm{Pa}\). ViWi more closely
matches the ground-truth magnitudes, with reduced local errors in the highlighted
regions.
}
    \vspace{-4pt}
    \label{fig:qualitative_comparison}
\end{figure*}

\subsection{Qualitative Results}
\label{sec:qualitative_results}

% Figure~\ref{fig:qualitative_comparison} compares ViWi with
% VoMP~\citep{dagli2025vomp} and the ground truth on two representative GVM
% objects. ViWi produces volumetric property fields that more closely match the
% ground-truth spatial distributions. The improvement is particularly pronounced for
% Young's modulus, where VoMP assigns incorrect stiffness values over large regions,
% whereas ViWi better recovers both the overall magnitude and local material
% structure. ViWi also provides more accurate and spatially coherent estimates of
% Poisson's ratio and density.
% % The highlighted regions further show that ViWi reduces local voxel-wise prediction
% % errors. For the \textit{sofa}, the relative error in Young's modulus decreases from
% % \(140.7\%\) to \(7.7\%\), while for the \textit{table} it decreases from \(49.3\%\)
% % to \(6.0\%\). Similar qualitative improvements are visible across the remaining
% % properties and examples. These results further support the benefit of combining
% % material-slot grouping with complementary RF evidence for coherent volumetric
% % mechanical-property estimation.
% The highlighted regions further show lower local voxel-wise errors. For the
% \textit{sofa} and \textit{table}, the relative error in Young's modulus decreases
% from \(140.7\%\) to \(7.7\%\) and from \(49.3\%\) to \(6.0\%\), respectively.
% % Similar improvements are visible across the remaining properties and examples,
% % supporting the benefit of material-slot grouping with complementary RF evidence.

Figure~\ref{fig:qualitative_comparison} compares ViWi with VoMP~\citep{dagli2025vomp}
and the ground truth on two representative GVM test objects. VoMP's dominant failure is
one of magnitude: it assigns large regions the wrong stiffness level, over-stiffening
the sofa and under-stiffening the table, whereas ViWi recovers the correct level for the
dominant material of each object. Averaged over all voxels, the relative error in \(E\)
(in linear \(\mathrm{Pa}\)) falls from \(140.7\%\) to \(7.7\%\) on the sofa and from
\(49.3\%\) to \(6.0\%\) on the table, with consistent gains for \(\nu\) and \(\rho\); the highlighted regions show
the same discrepancy at the voxel scale.

\begin{wraptable}{r}{0.49\textwidth}
\vspace{-36pt}
\centering
\scriptsize
\setlength{\tabcolsep}{10pt}

\caption{\textbf{Computational cost.} Inference at batch size~1.}
\label{tab:cost}
\vspace{-3pt}

\begin{tabular}{@{}lrrr@{}}
\toprule
 & VoMP & ViWi$^{\dagger}$ & ViWi \\
\midrule
\multicolumn{4}{@{}l}{\emph{Training} (to 112k steps)} \\
\quad Trainable params     & --   & 1.34\,M & 1.88\,M \\
% \quad GPUs (H100-80GB)     & --   & 4       & 8 \\
\quad Number of GPUs     & --   & 4       & 8 \\
\quad Time [h]             & --   & 24.5    & 32.6 \\
\quad Memory [GB/GPU]      & --   & 7.0     & 7.0 \\
\midrule
\multicolumn{4}{@{}l}{\emph{Inference} (per object)} \\
\quad 4.3k voxels [ms]     &  8.4 &  9.9 & 10.3 \\
\quad 28.8k voxels [ms]    & 25.4 & 27.6 & 27.9 \\
\quad 32.8k voxels [ms]    & 28.2 & 30.3 & 30.7 \\
\quad Mean, 32 obj. [ms]   & 24.3 & 26.3 & 26.6 \\
\quad Memory [MB]          &  758 &  763 &  765 \\
\bottomrule
\end{tabular}

\vspace{-8pt}
\end{wraptable}

\subsection{Computational Cost}

ViWi adds a lightweight material-slot head on top of the frozen VoMP geometry
backbone and MatVAE, introducing only \(1.88\)M trainable parameters. As shown in
Table~\ref{tab:cost}, the additional computation and memory overhead at inference
are modest, and RF conditioning adds only a small cost over
ViWi\(^{\dagger}\). Training remains practical, while RF descriptors are computed
once offline. Overall, the computational overhead of ViWi is small relative to the
underlying volumetric backbone.

\section{Conclusion and Limitations}
\label{sec:conclusion}

We introduced ViWi, a multimodal framework for volumetric mechanical-property
estimation that combines object-centric material-slot decomposition with
complementary RF evidence. ViWi groups voxels according to visual and
material-latent compatibility, producing coherent slot-level property prototypes
rather than independent voxel-wise predictions. A physics-based RF descriptor
further conditions the material slots and helps resolve cases in which visual
appearance alone is insufficient to distinguish underlying materials.

ViWi improves four of the six reported metrics on GVM, with particularly strong
gains for Young's modulus and Poisson's ratio. Its vision-only variant also improves
real-object mass estimation on ABO-500, showing that the material-slot representation
generalizes beyond the synthetic volumetric benchmark. 
 % Ablation studies further show
% that RF is most beneficial when visual evidence is less reliable and remains robust
% to moderate descriptor perturbations.

A key limitation is that the RF descriptors are simulated rather than measured
with real hardware and therefore do not capture all sources of variation encountered
in practical RF sensing. Future work will evaluate ViWi using real RF observations
under realistic sensing conditions.

\bibliographystyle{unsrtnat}
\bibliography{references}  %%% Uncomment this line and comment out the ``thebibliography'' section below to use the external .bib file (using bibtex) .





% \end{thebibliography}

%%%%%%%%%%%%%%%%%%%%%%%%%%%%%%%%%%%%%%%%%%%%%%%%%%%%%%%%%%%%
\newpage

\section*{\centering \textcolor{darkgreen}{Vi}sion Meets \textcolor{darkgreen}{Wi}Fi \\
Physics-Grounded Estimation of Volumetric Material Properties \\ Supplementary Material}
\vspace{1cm}

\appendix

\section{Experimental Setup}

\paragraph{Datasets.}
We evaluate ViWi on the same two datasets and under the same evaluation protocol as
VoMP~\citep{dagli2025vomp}. The primary evaluation is conducted on GVM, a collection
of simulation-ready 3D assets represented on a \(64^3\) voxel grid. Each object is
segmented into material regions, and every occupied voxel is assigned ground-truth
Young's modulus \(E\), Poisson's ratio \(\nu\), and density \(\rho\) through a
segment-to-material-to-property annotation pipeline. We use the predefined
object-level training, validation, and test splits and report results on the
held-out test set. As 18 of the 166 test objects are absent from the public GVM
release, we evaluate on the 148 released objects and re-run VoMP on this same
subset so that all reported numbers are directly comparable.

We additionally evaluate mass estimation on ABO-500, which contains real product
assets with measured object masses. Following VoMP, we use the 100-object test
subset only for evaluation and do not train or fine-tune on ABO-500. Since
voxel-level mechanical-property annotations are unavailable, RF descriptors cannot
be constructed for these objects; we therefore evaluate ViWi\(^{\dagger}\), our
vision-only variant. The predicted density field is integrated over the occupied
object volume and compared with the measured object mass. This evaluation assesses
whether the learned volumetric density fields transfer from simulated assets to
real objects.

\paragraph{RF descriptor simulation.}
Our RF descriptors are simulated with Sionna RT in three stages, all purely
additive: the original GVM voxel and feature files are never modified.

\emph{(i) Per-voxel electromagnetic properties.} Each voxel of an object carries a
free-form material string from the GVM annotations. We map every string to one of
\(12\) material categories by exact name match, then by an ordered set of keyword
rules (e.g.\ \texttt{chrome}\(\to\)metal, \texttt{pine\_bark}\(\to\)wood), and
finally, for uninformative names, by nearest-centroid assignment in
\((\log_{10}E,\ \nu,\ \log_{10}\rho)\) space, allowing the mechanical properties
to resolve ambiguous material names. Each category is assigned a relative
permittivity \(\varepsilon_r\) and conductivity \(\sigma\) at \(2.4\)~GHz, taken
from ITU-R P.2040 where the material is tabulated (wood, glass, concrete, ground)
and from the antenna-materials literature otherwise; metal is modeled as a
near-perfect conductor. The resulting per-voxel \((\varepsilon_r,\sigma)\) fields
are stored in the same row order as the voxel grid and therefore align
index-for-index with the mechanical ground truth.

\emph{(ii) Geometry.} Voxels are grouped by material category and converted to one
surface triangle mesh per category. Only \emph{exposed} faces are emitted, i.e.\
faces whose neighboring voxel is empty, yielding a thin surface shell rather than
a union of solid cubes and substantially reducing the triangle count for ray
tracing. Each object is centered at the origin and isotropically scaled so that its
largest bounding-box dimension equals \(1\)~m, making the sensing geometry
comparable across objects with different native scales.

\emph{(iii) Propagation.} Each category mesh is assigned its own
\texttt{RadioMaterial}\((\varepsilon_r,\sigma)\), and
\(N_{\mathrm{tx}}\) transmitters are placed on a Fibonacci sphere of radius
\(1.5\)~m around the object, with \(32\) receivers on a concentric sphere at
\(1.05\times\) that radius. All elements are isotropic and vertically polarized.
We trace paths up to depth \(5\) with specular reflection, diffuse reflection, and
refraction enabled, while disabling the line-of-sight path so that the descriptor
captures the object's scattering response rather than direct
transmitter--receiver coupling. For each transmitter, we aggregate the complex
path coefficients over all receivers and paths into four statistics: total
scattered power (dB), the fraction of receivers with a valid path, maximum
single-path gain (dB), and power-weighted RMS delay spread (ns). Concatenating
these statistics over transmitters gives a fixed
\(4N_{\mathrm{tx}}\)-dimensional RF descriptor per object
(\(64\) dimensions for \(N_{\mathrm{tx}}=16\)), independent of the voxel count
\(N\). Transmitters with no valid return are floored at \(-120\)~dB. Each
descriptor dimension is then z-scored using fixed normalization statistics shared
across the train, validation, and test splits.

\paragraph{Evaluation metrics.}
For volumetric mechanical-property estimation on GVM, we follow
VoMP~\citep{dagli2025vomp}. Errors are first computed over the occupied voxels of
each object and then averaged across objects to avoid weighting objects by their
number of occupied voxels. We report Average Log Displacement Error (ALDE) and
Average Log Relative Error (ALRE) for Young's modulus, and Average Displacement
Error (ADE) and Average Relative Error (ARE) for Poisson's ratio and density.

For ABO-500, we evaluate the mass obtained from the predicted volumetric density
field and report ALDE, ADE, ARE, and minimum ratio error (MnRE), following VoMP.
Lower values are better for ALDE, ADE, and ARE, whereas higher MnRE indicates better
agreement between predicted and ground-truth masses.

\paragraph{Implementation details.}
Unless otherwise specified, ViWi uses \(K=16\) material slots with embedding
dimension \(d_s=256\), refined for \(T=3\) slot refinement iterations. The RF
descriptor contains four statistics for each of sixteen transmitters, resulting in
a \(64\)-dimensional input. We use the modulation bound \(c=0.3\) and set
\(\lambda_{\mathrm{mat}}=\lambda_{\mathrm{comp}}=0.1\). The pretrained geometry
backbone and MatVAE are kept frozen.

\section{Additional Ablation Studies}

\paragraph{Effect of additional vision-only training (Ablations I--II).}
\label{sec:ablation_more_training}

Because ViWi is initialized from ViWi\(^{\dagger}\), our vision-only variant, its
improvement could partly result from additional optimization rather than RF
conditioning. We therefore continue training ViWi\(^{\dagger}\) for the same number
of steps while keeping RF disabled. Table~\ref{tab:combined_ablation} reports this
control on both the full test set (Ablation II) and the visually ambiguous subset
(Ablation I), where mean error denotes the average relative error across
\(\log_{10}E\), \(\nu\), and \(\rho\). On the full set, extended vision-only
training reduces the mean error from \(6.50\%\) to \(6.10\%\), whereas ViWi further
reduces it to \(5.81\%\). The same trend is stronger on visually ambiguous objects,
where the error decreases from \(7.15\%\) to \(6.79\%\) with additional vision-only
training and to \(6.21\%\) with RF. These results show that the gains of ViWi cannot
be explained by additional training alone and that RF provides complementary
information, particularly when visual evidence is less reliable.

\paragraph{Slot refinement iterations (Ablation III).}
\label{sec:ablation_slot_iterations}

As shown in Table~\ref{tab:combined_ablation}, increasing the number of slot refinement iterations from \(T=3\) to \(T=7\)
redistributes accuracy across properties rather than improving it uniformly.
Young's modulus and Poisson's ratio degrade (ALDE
\(0.185\!\to\!0.230\), \(+24\%\); ADE
\(0.0146\!\to\!0.0167\), \(+14\%\)), while density improves
(ADE \(238.7\!\to\!201.1\), \(-16\%\); ARE
\(0.124\!\to\!0.104\), \(-17\%\)). Because the mean error averages the relative
errors across \(\log_{10}E\), \(\nu\), and \(\rho\), the density improvement
reduces the aggregate error from \(6.50\%\) to \(6.20\%\). We retain \(T=3\) as the default because it requires less than half the refinement iterations.

\paragraph{RF noise robustness (Ablation IV).}
We evaluate robustness by adding zero-mean Gaussian noise to the standardized RF
descriptor at inference while keeping the model fixed. As shown in
Table~\ref{tab:combined_ablation}, performance remains stable under mild
perturbations: the mean error changes from \(5.82\%\) with clean RF to
\(5.81\%\) at \(\sigma=0.1\) and \(6.02\%\) at \(\sigma=0.5\). Even under the
strong perturbation \(\sigma=1.0\), the error increases only to \(6.28\%\).
These results show that ViWi degrades gradually as RF corruption increases and is
robust to moderate perturbations of the RF descriptor.

\begin{table}[t]
\centering
\caption{
{Additional ablation studies on GVM.}
Results for training duration, slot refinement, RF noise robustness, unseen materials,
and number of RF transmitters. Metrics are first averaged over occupied voxels within
each object and then averaged across objects. Mean error is the average relative error
across \(\log_{10}E\), \(\nu\), and \(\rho\). Lower is better for all metrics.
}
\label{tab:combined_ablation}
\vspace{4pt}
\footnotesize

\resizebox{\textwidth}{!}{%
\begin{tabular}{@{}l *{7}{c} @{}}
\toprule
\multirow{2}{*}{Configuration}
& \multicolumn{2}{c}{Young's Modulus (\(E\))}
& \multicolumn{2}{c}{Poisson's Ratio (\(\nu\))}
& \multicolumn{2}{c}{Density (\(\rho\))}
& \multirow{2}{*}{Mean Error (\%)} \\
\cmidrule(lr){2-3}
\cmidrule(lr){4-5}
\cmidrule(lr){6-7}
& ALDE (\(\downarrow\))
& ALRE (\(\downarrow\))
& ADE (\(\downarrow\))
& ARE (\(\downarrow\))
& ADE (\(\downarrow\))
& ARE (\(\downarrow\))
& \\
\midrule

\midrule
\multicolumn{8}{@{}l}{\textit{Ablation I: Extended Vision-Only training,
visually ambiguous objects (80 of 148)}} \\
\addlinespace[2pt]

ViWi\(^{\dagger}\)
& {0.2085 {\scriptsize (\(\pm\)0.2036)}}
& {0.0232 {\scriptsize (\(\pm\)0.0259)}}
& {0.0138 {\scriptsize (\(\pm\)0.0140)}}
& {0.0483 {\scriptsize (\(\pm\)0.0533)}}
& {244.51 {\scriptsize (\(\pm\)234.6938)}}
& {0.1430 {\scriptsize (\(\pm\)0.1274)}}
& {7.15 {\scriptsize (\(\pm\)0.0517)}} \\

ViWi\(^{\dagger}\) (extended training)
& {0.2132 {\scriptsize (\(\pm\)0.2514)}}
& {0.0240 {\scriptsize (\(\pm\)0.0312)}}
& {0.0136 {\scriptsize (\(\pm\)0.0150)}}
& {0.0468 {\scriptsize (\(\pm\)0.0518)}}
& {211.83 {\scriptsize (\(\pm\)194.4298)}}
& {0.1327 {\scriptsize (\(\pm\)0.1343)}}
& {6.79 {\scriptsize (\(\pm\)0.0574)}} \\

ViWi
& {0.2052 {\scriptsize (\(\pm\)0.2196)}}
& {0.0227 {\scriptsize (\(\pm\)0.0270)}}
& {0.0148 {\scriptsize (\(\pm\)0.0155)}}
& {0.0512 {\scriptsize (\(\pm\)0.0557)}}
& {194.31 {\scriptsize (\(\pm\)211.8055)}}
& {0.1123 {\scriptsize (\(\pm\)0.0995)}}
& {6.21 {\scriptsize (\(\pm\)0.0474)}} \\

\midrule
\multicolumn{8}{@{}l}{\textit{Ablation II: Extended Vision-Only training}} \\
\addlinespace[2pt]

ViWi\(^{\dagger}\)
& {0.1853 {\scriptsize (\(\pm\)0.2026)}}
& {0.0202 {\scriptsize (\(\pm\)0.0249)}}
& {0.0146 {\scriptsize (\(\pm\)0.0151)}}
& {0.0503 {\scriptsize (\(\pm\)0.0558)}}
& {238.71 {\scriptsize (\(\pm\)304.1573)}}
& {0.1243 {\scriptsize (\(\pm\)0.1198)}} 
& {6.50 {\scriptsize (\(\pm\)0.0492)}} \\

ViWi\(^{\dagger}\) (extended training)
& {0.1812 {\scriptsize (\(\pm\)0.2271)}}
& {0.0200 {\scriptsize (\(\pm\)0.0276)}}
& {0.0145 {\scriptsize (\(\pm\)0.0152)}}
& {0.0491 {\scriptsize (\(\pm\)0.0504)}}
& {197.00 {\scriptsize (\(\pm\)238.3770)}}
& {0.1140 {\scriptsize (\(\pm\)0.1228)}} 
& {6.10 {\scriptsize (\(\pm\)0.0516)}} \\

ViWi
& {0.1797 {\scriptsize (\(\pm\)0.2121)}}
& {0.0196 {\scriptsize (\(\pm\)0.0257)}}
& {0.0151 {\scriptsize (\(\pm\)0.0152)}}
& {0.0515 {\scriptsize (\(\pm\)0.0516)}}
& {198.22 {\scriptsize (\(\pm\)340.8454)}}
& {0.1034 {\scriptsize (\(\pm\)0.1080)}} 
& {5.81 {\scriptsize (\(\pm\)0.0475)}} \\

\midrule
\multicolumn{8}{@{}l}{\textit{Ablation III: Slot refinement iterations}} \\
\addlinespace[2pt]

ViWi\(^{\dagger}\), \(T=3\)
& {0.1853 {\scriptsize (\(\pm\)0.2026)}}
& {0.0202 {\scriptsize (\(\pm\)0.0249)}}
& {0.0146 {\scriptsize (\(\pm\)0.0151)}}
& {0.0503 {\scriptsize (\(\pm\)0.0558)}}
& {238.71 {\scriptsize (\(\pm\)304.1573)}}
& {0.1243 {\scriptsize (\(\pm\)0.1198)}} 
& {6.50 {\scriptsize (\(\pm\)0.0492)}} \\

ViWi\(^{\dagger}\), \(T=7\)
& {0.2304 {\scriptsize (\(\pm\)0.2366)}}
& {0.0246 {\scriptsize (\(\pm\)0.0271)}}
& {0.0167 {\scriptsize (\(\pm\)0.0139)}}
& {0.0578 {\scriptsize (\(\pm\)0.0497)}}
& {201.12 {\scriptsize (\(\pm\)341.5979)}}
& {0.1038 {\scriptsize (\(\pm\)0.1138)}} 
& {6.20 {\scriptsize (\(\pm\)0.0458)}} \\

\midrule
\multicolumn{8}{@{}l}{\textit{Ablation IV: RF noise robustness}} \\
\addlinespace[2pt]

ViWi
& {0.1797 {\scriptsize (\(\pm\)0.2121)}}
& {0.0196 {\scriptsize (\(\pm\)0.0257)}}
& {0.0151 {\scriptsize (\(\pm\)0.0152)}}
& {0.0515 {\scriptsize (\(\pm\)0.0516)}}
& {198.22 {\scriptsize (\(\pm\)340.8454)}}
& {0.1034 {\scriptsize (\(\pm\)0.1080)}} 
& {5.81 {\scriptsize (\(\pm\)0.0475)}} \\

ViWi, \(\sigma=0.10\)
& {0.1812 {\scriptsize (\(\pm\)0.2138)}}
& {0.0197 {\scriptsize (\(\pm\)0.0257)}}
& {0.0151 {\scriptsize (\(\pm\)0.0152)}}
& {0.0515 {\scriptsize (\(\pm\)0.0516)}}
& {198.22 {\scriptsize (\(\pm\)340.8454)}}
& {0.1034 {\scriptsize (\(\pm\)0.1080)}} 
& {5.81 {\scriptsize (\(\pm\)0.0475)}} \\

ViWi, \(\sigma=0.25\)
& {0.1841 {\scriptsize (\(\pm\)0.2091)}}
& {0.0200 {\scriptsize (\(\pm\)0.0251)}}
& {0.0156 {\scriptsize (\(\pm\)0.0152)}}
& {0.0532 {\scriptsize (\(\pm\)0.0519)}}
& {208.05 {\scriptsize (\(\pm\)391.1836)}}
& {0.1045 {\scriptsize (\(\pm\)0.1082)}} 
& {5.92 {\scriptsize (\(\pm\)0.0473)}} \\

ViWi, \(\sigma=0.50\)
& {0.1838 {\scriptsize (\(\pm\)0.2046)}}
& {0.0200 {\scriptsize (\(\pm\)0.0246)}}
& {0.0158 {\scriptsize (\(\pm\)0.0151)}}
& {0.0536 {\scriptsize (\(\pm\)0.0522)}}
& {213.47 {\scriptsize (\(\pm\)388.0237)}}
& {0.1071 {\scriptsize (\(\pm\)0.1062)}} 
& {6.02 {\scriptsize (\(\pm\)0.0467)}} \\

ViWi, \(\sigma=1.00\)
& {0.1986 {\scriptsize (\(\pm\)0.2120)}}
& {0.0218 {\scriptsize (\(\pm\)0.0259)}}
& {0.0154 {\scriptsize (\(\pm\)0.0145)}}
& {0.0520 {\scriptsize (\(\pm\)0.0510)}}
& {224.55 {\scriptsize (\(\pm\)393.4308)}}
& {0.1145 {\scriptsize (\(\pm\)0.1062)}} 
& {6.28 {\scriptsize (\(\pm\)0.0463)}} \\

\midrule
\multicolumn{8}{@{}l}{\textit{Ablation V: Unseen material combination}} \\
\addlinespace[2pt]

ViWi\(^{\dagger}\)
& {0.7002 {\scriptsize (\(\pm\)0.5513)}}
& {0.0784 {\scriptsize (\(\pm\)0.0701)}}
& {0.0343 {\scriptsize (\(\pm\)0.0340)}}
& {0.1164 {\scriptsize (\(\pm\)0.1293)}}
& {887.01 {\scriptsize (\(\pm\)1004.8889)}}
& {0.4192 {\scriptsize (\(\pm\)0.5558)}} 
& {20.47 {\scriptsize (\(\pm\)0.2030)}} \\

ViWi
& {0.7232 {\scriptsize (\(\pm\)0.5160)}}
& {0.0756 {\scriptsize (\(\pm\)0.0529)}}
& {0.0318 {\scriptsize (\(\pm\)0.0271)}}
& {0.1059 {\scriptsize (\(\pm\)0.1018)}}
& {757.73 {\scriptsize (\(\pm\)838.1707)}}
& {0.3040 {\scriptsize (\(\pm\)0.3651)}} 
& {16.18 {\scriptsize (\(\pm\)0.1341)}} \\

\midrule
\multicolumn{8}{@{}l}{\textit{Ablation VI: Number of RF transmitters}} \\
\addlinespace[2pt]

% ViWi, \(N_{\mathrm{tx}}=4\)
% & {0.1935 {\scriptsize (\(\pm\)0.2215)}}
% & {0.0208 {\scriptsize (\(\pm\)0.0260)}}
% & {0.0150 {\scriptsize (\(\pm\)0.0146)}}
% & {0.0525 {\scriptsize (\(\pm\)0.0541)}}
% & {229.32 {\scriptsize (\(\pm\)321.4760)}}
% & {0.1161 {\scriptsize (\(\pm\)0.1249)}} 
% & {6.31 {\scriptsize (\(\pm\)0.0533)}} \\

% ViWi, \(N_{\mathrm{tx}}=16\)
% & {0.1797 {\scriptsize (\(\pm\)0.2121)}}
% & {0.0196 {\scriptsize (\(\pm\)0.0257)}}
% & {0.0151 {\scriptsize (\(\pm\)0.0152)}}
% & {0.0515 {\scriptsize (\(\pm\)0.0516)}}
% & {198.22 {\scriptsize (\(\pm\)340.8454)}}
% & {0.1034 {\scriptsize (\(\pm\)0.1080)}} 
% & {5.81 {\scriptsize (\(\pm\)0.0475)}} \\

ViWi, \(N_{\mathrm{tx}}=4\)
& {0.1934 {\scriptsize (\(\pm\)0.2204)}}
& {0.0207 {\scriptsize (\(\pm\)0.0258)}}
& {0.0157 {\scriptsize (\(\pm\)0.0145)}}
& {0.0549 {\scriptsize (\(\pm\)0.0529)}}
& {230.33 {\scriptsize (\(\pm\)353.14)}}
& {0.1180 {\scriptsize (\(\pm\)0.1337)}} 
& {6.45 {\scriptsize (\(\pm\)0.0560)}} \\

ViWi, \(N_{\mathrm{tx}}=16\)
& {0.1951 {\scriptsize (\(\pm\)0.2403)}}
& {0.0213 {\scriptsize (\(\pm\)0.0287)}}
& {0.0164 {\scriptsize (\(\pm\)0.0159)}}
& {0.0569 {\scriptsize (\(\pm\)0.0547)}}
& {190.92 {\scriptsize (\(\pm\)246.10)}}
& {0.0980 {\scriptsize (\(\pm\)0.0995)}} 
& {5.87 {\scriptsize (\(\pm\)0.0452)}} \\

\midrule
\multicolumn{8}{@{}l}{\textit{Ablation VII: Tx/Rx sphere radius}} \\
\addlinespace[2pt]

% ViWi, \(r=1.0\,\mathrm{m}\)
% & {0.2108 {\scriptsize (\(\pm\)0.2038)}}
% & {0.0231 {\scriptsize (\(\pm\)0.0250)}}
% & {0.0167 {\scriptsize (\(\pm\)0.0150)}}
% & {0.0589 {\scriptsize (\(\pm\)0.0566)}}
% & {199.96 {\scriptsize (\(\pm\)286.92)}}
% & {0.1068 {\scriptsize (\(\pm\)0.1044)}}
% & {6.29 {\scriptsize (\(\pm\)0.0476)}} \\

ViWi, \(r=1.5\,\mathrm{m}\) (train)
& {{0.1797} {\scriptsize (\(\pm\)0.2121)}}
& {{0.0196} {\scriptsize (\(\pm\)0.0257)}}
& {0.0151 {\scriptsize (\(\pm\)0.0152)}}
& {0.0515 {\scriptsize (\(\pm\)0.0516)}}
& {198.22 {\scriptsize (\(\pm\)340.85)}}
& {{0.1034} {\scriptsize (\(\pm\)0.1080)}}
& {5.81 {\scriptsize (\(\pm\)0.0475)}} \\

ViWi, \(r=2.0\,\mathrm{m}\)
& {0.1877 {\scriptsize (\(\pm\)0.2193)}}
& {0.0206 {\scriptsize (\(\pm\)0.0268)}}
& {{0.0140} {\scriptsize (\(\pm\)0.0141)}}
& {{0.0470} {\scriptsize (\(\pm\)0.0483)}}
& {{198.08} {\scriptsize (\(\pm\)297.99)}}
& {0.1063 {\scriptsize (\(\pm\)0.1092)}}
& {{5.80} {\scriptsize (\(\pm\)0.0479)}} \\

ViWi, \(r=4.0\,\mathrm{m}\)
& {0.2067 {\scriptsize (\(\pm\)0.2109)}}
& {0.0227 {\scriptsize (\(\pm\)0.0259)}}
& {0.0146 {\scriptsize (\(\pm\)0.0149)}}
& {0.0489 {\scriptsize (\(\pm\)0.0514)}}
& {213.06 {\scriptsize (\(\pm\)356.55)}}
& {0.1150 {\scriptsize (\(\pm\)0.1038)}}
& {6.22 {\scriptsize (\(\pm\)0.0462)}} \\

ViWi, \(r=8.0\,\mathrm{m}\)
& {0.2214 {\scriptsize (\(\pm\)0.2072)}}
& {0.0241 {\scriptsize (\(\pm\)0.0251)}}
& {0.0160 {\scriptsize (\(\pm\)0.0132)}}
& {0.0534 {\scriptsize (\(\pm\)0.0462)}}
& {216.98 {\scriptsize (\(\pm\)303.85)}}
& {0.1228 {\scriptsize (\(\pm\)0.1014)}}
& {6.68 {\scriptsize (\(\pm\)0.0450)}} \\

% ViWi, no RF
% & {0.1971 {\scriptsize (\(\pm\)0.2358)}}
% & {0.0214 {\scriptsize (\(\pm\)0.0281)}}
% & {0.0159 {\scriptsize (\(\pm\)0.0166)}}
% & {0.0554 {\scriptsize (\(\pm\)0.0588)}}
% & {200.89 {\scriptsize (\(\pm\)334.48)}}
% & {0.1063 {\scriptsize (\(\pm\)0.1197)}}
% & {6.10 {\scriptsize (\(\pm\)0.0548)}} \\

% ViWi, \(N_{\mathrm{tx}}=32\)
% & {0.1748 {\scriptsize (\(\pm\)0.2021)}}
% & {0.0190 {\scriptsize (\(\pm\)0.0242)}}
% & {0.0156 {\scriptsize (\(\pm\)0.0145)}}
% & {0.0535 {\scriptsize (\(\pm\)0.0493)}}
% & {214.27 {\scriptsize (\(\pm\)316.8948)}}
% & {0.1150 {\scriptsize (\(\pm\)0.1103)}} 
% & {6.25 {\scriptsize (\(\pm\)0.0432)}} \\

\bottomrule
\end{tabular}%
}
\end{table}

\paragraph{Generalization to unseen materials (Ablation V).}
\label{sec:ablation_unseen_materials}

Because the simulated RF descriptor is derived from material-dependent
electromagnetic properties, a possible concern is that the model may simply
memorize RF patterns associated with materials seen during training. To test this,
we hold out two material categories entirely. Using a per-object membership index
derived from the per-voxel material labels, we remove every object that contains
metal or rubber in \emph{any} voxel from the training and validation splits, which
reduces training from \(1255\) to \(505\) objects and validation from \(157\) to
\(69\). Evaluation is then performed \emph{only} on the \(61\) of \(148\) test
objects that contain at least one of the two held-out materials; the remaining
\(87\) test objects are not scored in this ablation, so the errors here are not
comparable to those of the full-test-split rows above. The holdout is defined at
the material-category level rather than by raw material names, so all variants are
excluded together (e.g.\ chrome, steel, brass, and aluminium are all mapped to
metal). The model therefore never observes these material categories during
training or model selection.

As shown in Table~\ref{tab:combined_ablation}, ViWi\(^{\dagger}\), our vision-only
variant, obtains a mean error of \(20.47\%\), while RF conditioning, warm-started
from the vision-only model, reduces it to \(16.18\%\), an improvement of \(4.29\)
percentage points (\(21\%\) relative). RF improves both density metrics (ADE
\(887.0\!\to\!757.7\), ARE \(0.419\!\to\!0.304\)) and three of the four metrics
for Young's modulus and Poisson's ratio. These results indicate that the RF
contribution is not explained solely by memorizing materials observed during
training and that the RF descriptor provides transferable material-sensitive
information for unseen material categories.

\paragraph{Number of RF transmitters (Ablation VI).}
As shown in Table~\ref{tab:combined_ablation}, Ablation~VI varies the number of transmitters \(N_{\mathrm{tx}}\) used to construct
the RF descriptor. Since each transmitter contributes four statistics,
\(N_{\mathrm{tx}}=4\) and \(16\) correspond to descriptor dimensions of \(16\)
and \(64\), respectively. Increasing the number of transmitters from \(4\) to \(16\)
reduces the mean error from \(6.31\%\) to \(5.81\%\), with improvements in five of
the six reported metrics. This suggests that broader angular coverage provides more
informative RF measurements. We therefore use \(N_{\mathrm{tx}}=16\) throughout.

\begin{table}[b]
\centering
\caption{
{Effect of the material-aware assignment score.}
Comparison of ViWi\(^{\dagger}\) with and without the material-compatibility term.
All metrics are computed globally over all occupied voxels, rather than per object
followed by averaging across objects. Mean error is the average relative error across
\(\log_{10}E\), \(\nu\), and \(\rho\). Lower is better for all metrics.
}
\label{tab:material_score_ablation}
\vspace{2pt}

\scriptsize
\setlength{\tabcolsep}{2.5pt}
\begin{tabular}{@{}lccccccc@{}}
\toprule
Configuration
& \multicolumn{2}{c}{Young's Modulus (\(E\))}
& \multicolumn{2}{c}{Poisson's Ratio (\(\nu\))}
& \multicolumn{2}{c}{Density (\(\rho\))}
& Mean Error (\%) \\
\cmidrule(lr){2-3}
\cmidrule(lr){4-5}
\cmidrule(lr){6-7}
&
ALDE
& ALRE
& ADE
& ARE
& ADE
& ARE
& \\
\midrule

ViWi\(^{\dagger}\), without material score
& 0.1969
& 0.0223
& 0.0156
& 0.0543
& 143.86
& 0.1061
& 6.09 \\

ViWi\(^{\dagger}\), with material score
& \textbf{0.1763}
& \textbf{0.0199}
& \textbf{0.0146}
& \textbf{0.0511}
& 164.45
& \textbf{0.0958}
& \textbf{5.56} \\

\bottomrule
\end{tabular}
\end{table}

\paragraph{Sensitivity to the sensing distance (Ablation VII).}
\label{sec:ablation_radius}

The RF descriptor is simulated with the transmitters and receivers on concentric
spheres around the object, whose radius sets how far the antennas are from it.
Every object is rescaled to a maximum size of \(1\,\mathrm{m}\), so the sensing
distance is comparable across objects. The descriptor has the same dimensionality
at any radius, so a trained model can directly process a descriptor simulated at a
new radius. We therefore re-simulate the test set at three larger radii and evaluate
the \emph{same} trained weights on each, without retraining, while keeping the
normalization statistics fixed to those of the training radius.

As shown in Table~\ref{tab:combined_ablation}, performance remains stable near the
training radius: at \(2.0\,\mathrm{m}\), the mean error is \(5.80\%\), compared
with \(5.81\%\) at the training radius of \(1.5\,\mathrm{m}\), with slightly
better Poisson's ratio estimation. Accuracy then degrades with distance, reaching
\(6.22\%\) at \(4.0\,\mathrm{m}\) and \(6.68\%\) at \(8.0\,\mathrm{m}\).
As the antennas move farther from the object, the simulator finds no valid
propagation path for an increasing fraction of Tx--Rx links, from \(16\%\) at
\(1.5\,\mathrm{m}\) to \(96\%\) at \(8.0\,\mathrm{m}\). Consequently, the RF
descriptor contains progressively less object-dependent information at larger
sensing distances. These results show that ViWi is robust to moderate changes in
sensing distance around the training configuration, while performance degrades at
substantially larger distances.

\begin{figure}[t]
    \centering
    \includegraphics[width=\textwidth]{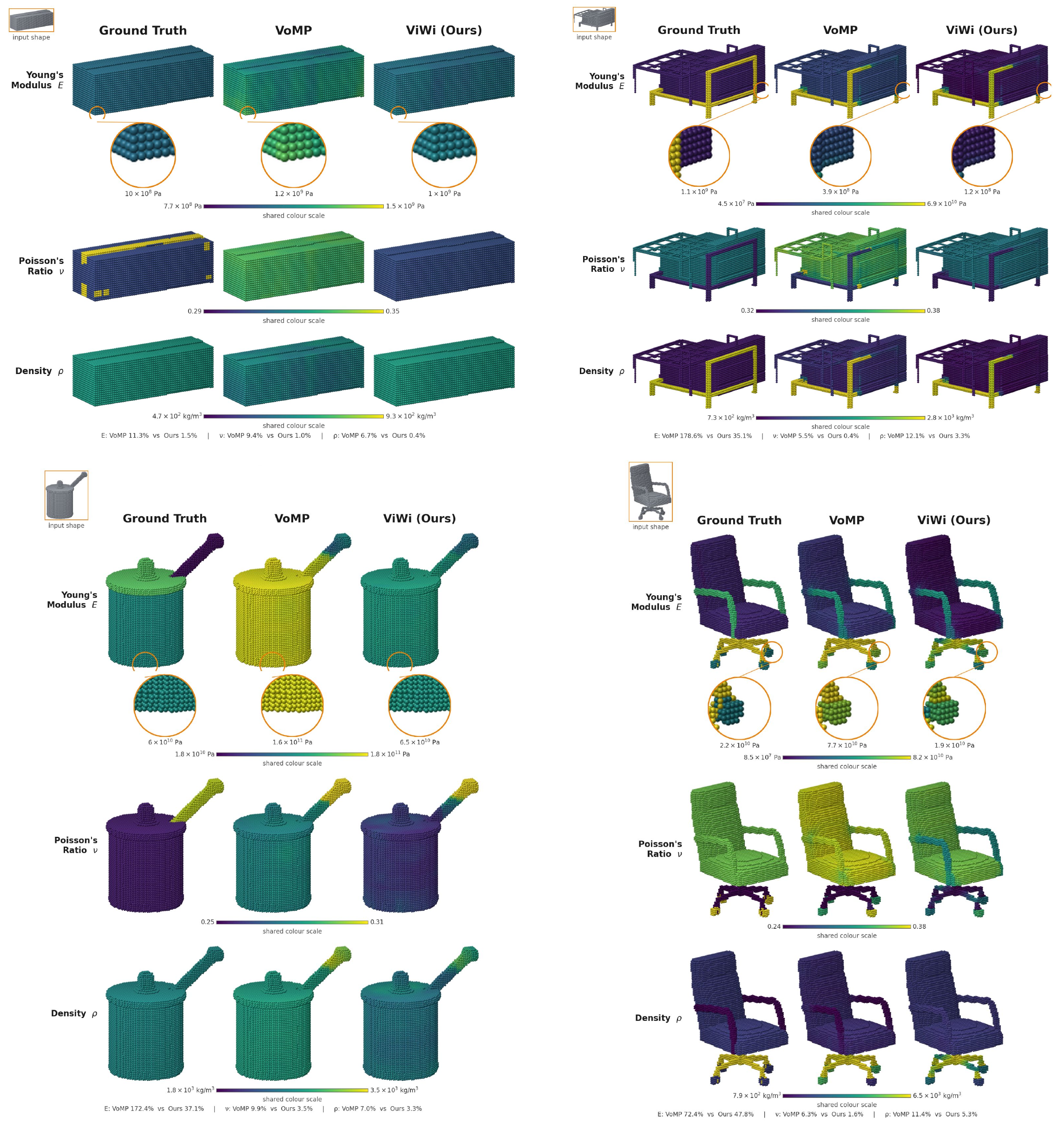}
    \caption{
    {Additional qualitative comparison of volumetric mechanical-property
    estimation.}
    Ground-truth fields are compared with predictions from
    VoMP~\citep{dagli2025vomp} and ViWi for Young's modulus \(E\),
    Poisson's ratio \(\nu\), and density \(\rho\).
    A shared color scale is used within each property and object.
    The magnified regions highlight local differences in Young's modulus predictions.
    }
    \label{fig:additional_qualitative_1}
\end{figure}

\paragraph{Contribution of the material-aware assignment score.}
\label{sec:ablation_material_score}

We evaluate the contribution of the material-compatibility term used in
voxel-to-slot assignment. In this ablation, we remove only the material score
\(a_{ik}^{\mathrm{mat}}\) from Eq.~\ref{eq:assignment_score}, so assignments are
determined solely by visual compatibility. The auxiliary material-latent loss
\(\mathcal{L}_{\mathrm{mat}}\) remains unchanged, isolating the effect of material
compatibility during slot assignment from that of material-latent supervision.

As shown in Table~\ref{tab:material_score_ablation}, removing the material score
increases the mean relative error from \(5.56\%\) to \(6.09\%\). Performance
degrades across all three properties: the relative error increases from
\(1.99\%\) to \(2.23\%\) for Young's modulus, from \(5.11\%\) to \(5.43\%\) for
Poisson's ratio, and from \(9.58\%\) to \(10.61\%\) for density. These results show
that visual compatibility alone is insufficient for reliable material grouping.
Compatibility in the learned material-latent space helps associate voxels with
material slots according to their inferred mechanical identity.

\section{Additional Qualitative Results}

\paragraph{Additional volumetric property visualizations.}
Figure~\ref{fig:additional_qualitative_1} provides additional qualitative comparisons
between ViWi, VoMP~\citep{dagli2025vomp}, and the ground truth on representative
GVM objects. Across objects with different geometries and material compositions,
ViWi more closely recovers the spatial distributions and magnitudes of Young's
modulus, Poisson's ratio, and density. The improvements are particularly visible in
regions where VoMP assigns incorrect mechanical-property values over spatially
coherent regions of the object. These additional examples further support the ability
of ViWi to produce more accurate and spatially consistent volumetric
mechanical-property fields.

\begin{figure}[t]
    \centering
    \includegraphics[width=\textwidth]{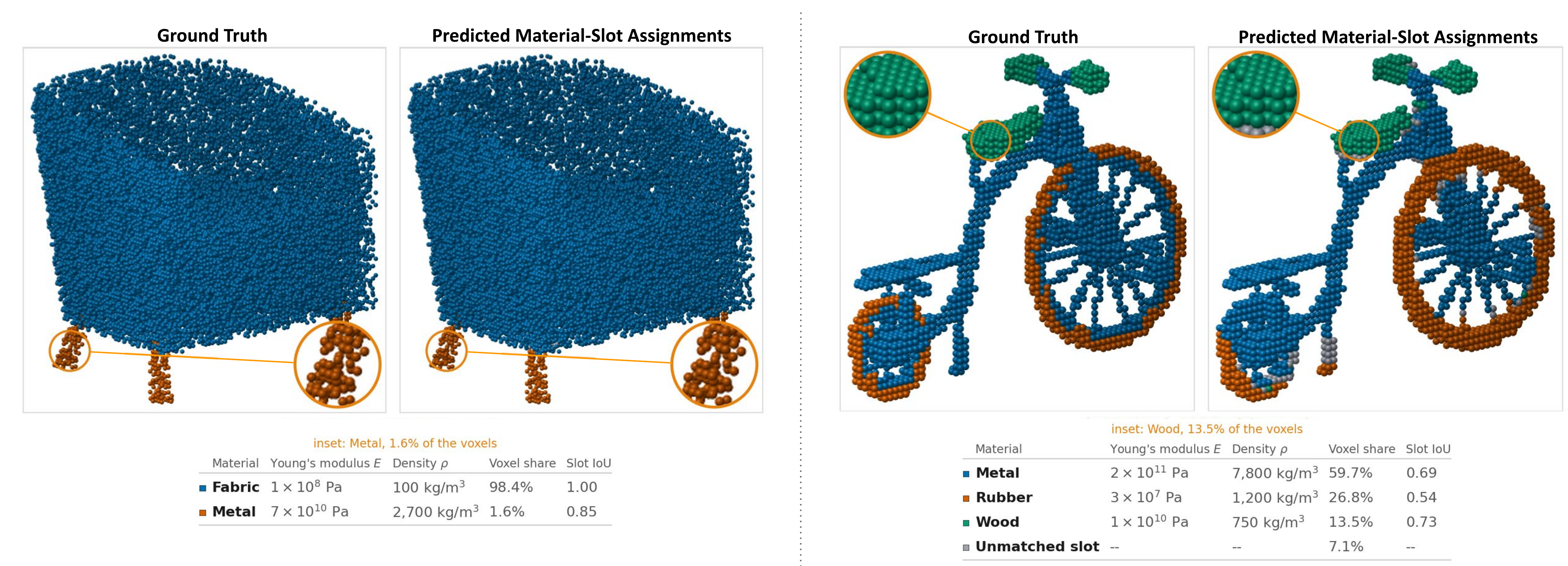}
    \caption{
    {Visualization of learned material-slot assignments.}
    Ground-truth material regions are compared with the material-slot assignments
    predicted by ViWi for an armchair (left) and a tricycle (right). Colors are
    matched between ground-truth materials and their corresponding predicted slots.
    The armchair slots recover the dominant fabric region and the small metal feet,
    while the tricycle slots separate metal, rubber, and wood components. Insets
    highlight small material regions. The tables report material properties, voxel
    shares, and IoU between each ground-truth material region and its matched slot.
    Gray denotes an unmatched predicted slot.
    }
    \label{fig:slots}
\end{figure}

\paragraph{What the material slots recover.}
Figure~\ref{fig:slots} compares the ground-truth material regions with the
material-slot assignments produced by ViWi\(^{\dagger}\). Each voxel is colored according to
its highest-weight voxel-to-slot assignment, with colors matched to the
corresponding ground-truth material regions. For the armchair (left), the dominant
slot recovers the fabric shell, which accounts for \(98.4\%\) of the occupied
voxels, while a second slot isolates the metal feet, which comprise only \(1.6\%\)
of the object and achieve an IoU of \(0.85\). Despite their small spatial extent,
the metal components differ substantially from the fabric in their mechanical
properties.

The tricycle (right) exhibits a more complex material composition. ViWi\(^{\dagger}\) separates
the metal structure (\(59.7\%\) of voxels), rubber components (\(26.8\%\)), and
wooden components (\(13.5\%\)), with slot IoUs of \(0.69\), \(0.54\), and
\(0.73\), respectively. Importantly, individual slots can group spatially
disconnected components that share a material identity, such as the separate
metal feet of the armchair and the wooden components of the tricycle. This behavior
is consistent with our formulation, where voxel-to-slot assignment is driven by
visual and material-latent compatibility rather than spatial proximity.

The assignments are not intended to provide exact material segmentation. In the
tricycle example, \(7.1\%\) of the voxels are assigned to a slot that does not
match any ground-truth material region, and some disagreement remains near part
boundaries. Nevertheless, the visualizations show that the learned material slots
recover meaningful material structure and aggregate voxels with a shared material
identity.

\paragraph{Physics simulation with predicted material fields.}
Accurate estimates of \((E,\nu,\rho)\) are most useful when they lead to more
accurate physical behavior. We therefore evaluate the predicted volumetric
mechanical-property fields inside a simulator using two complementary scenarios:
rigid-ball impact and quasi-static compression. In both cases, the same object is
simulated three times using the ground-truth mechanical-property field, the field
predicted by VoMP~\citep{dagli2025vomp}, and the field predicted by ViWi. All
rollouts share the same initial state and the same reduced basis, trained using a
uniform material field, so that the mechanical-property field is the only
prediction-specific quantity that changes between simulations. We measure the
deviation of each prediction-driven trajectory from the ground-truth-driven
trajectory using the mean pointwise distance over the simulated object, normalized
by the object's spatial extent. Lower values indicate closer agreement with the
reference simulation.

\begin{figure}[t]
    \centering
    \includegraphics[width=\textwidth]{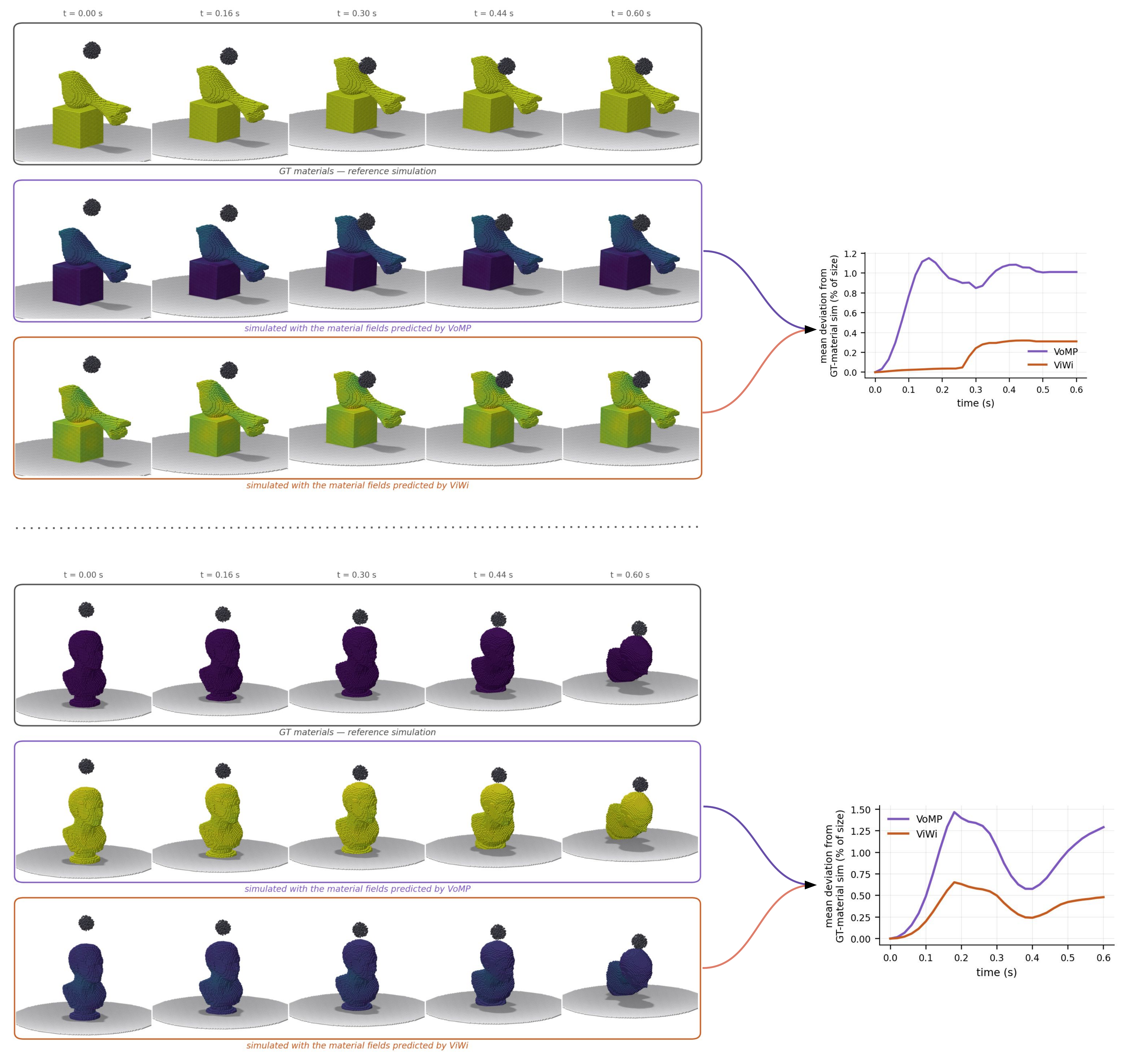}
    \caption{
    {Impact simulation using predicted material fields.}
    Two representative objects are simulated under identical rigid-ball impacts using
    the ground-truth material field, the field predicted by VoMP~\citep{dagli2025vomp},
    and the field predicted by ViWi. For each object, the left panel shows simulation
    snapshots over time, while the plot on the right reports the trajectory deviation
    from the ground-truth-driven simulation, normalized by object size. ViWi remains
    consistently closer to the reference simulation after contact.
    }
    \label{fig:impact}
\end{figure}

\paragraph{Impact simulation.}
We first consider a localized dynamic interaction in which a rigid ball is
released above the object and allowed to strike it. Figure~\ref{fig:impact}
shows two representative examples. For each object, the filmstrips compare the
ground-truth-, VoMP-, and ViWi-driven simulations under identical impact
conditions, while the plot on the right reports the deviation from the
ground-truth-driven trajectory over time. Before contact, the trajectories remain
nearly identical because the object has not yet responded to the applied impact.
After contact, the trajectories diverge according to the predicted
mechanical-property fields. In both examples, the simulation driven by ViWi remains
consistently closer to the reference trajectory than the simulation driven by VoMP.
These results show that improved volumetric mechanical-property estimation can
translate into more faithful downstream physical dynamics under localized impact.

\begin{figure}[t]
    \centering
    \includegraphics[width=\textwidth]{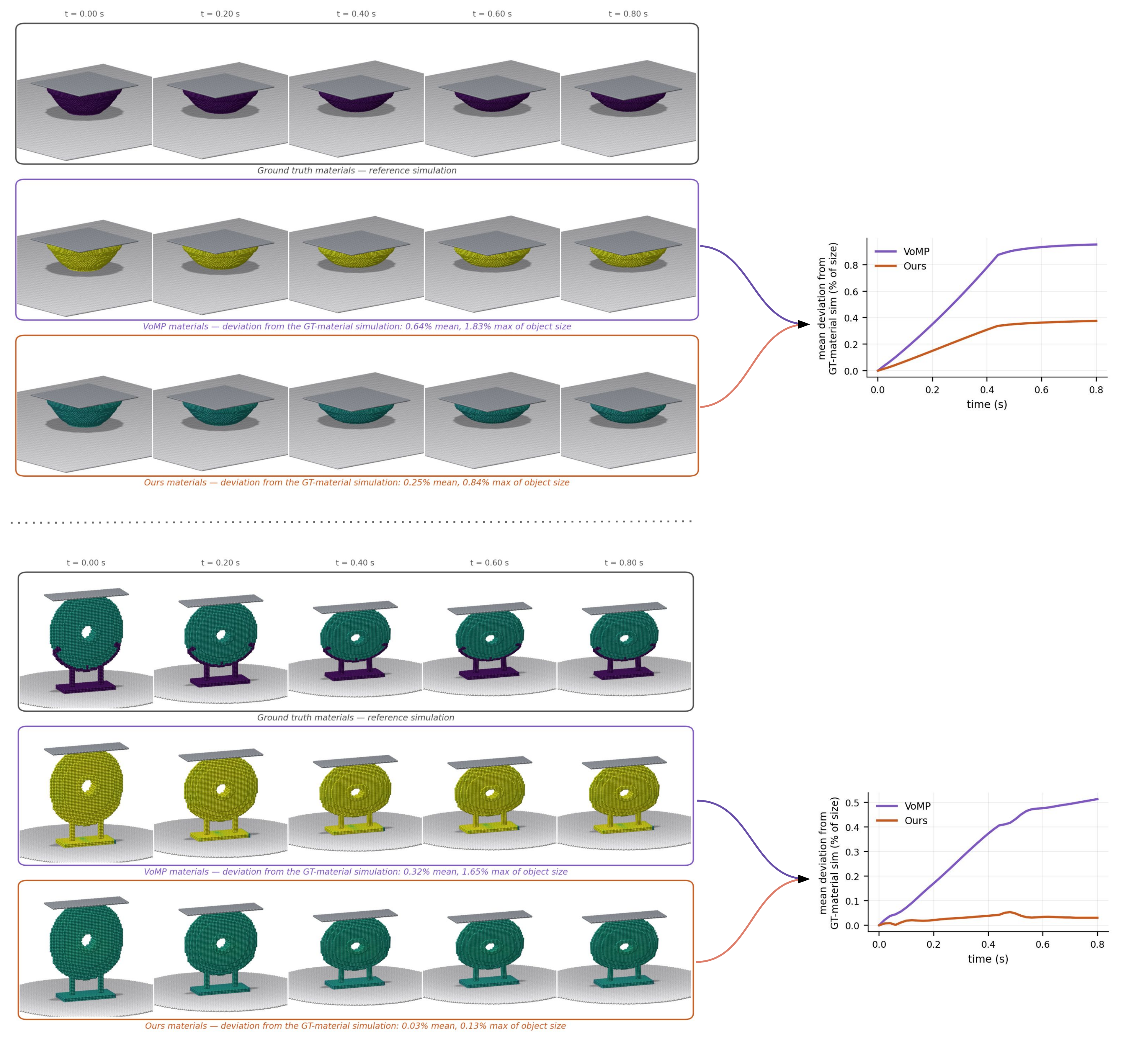}
    \caption{
    {Quasi-static compression using predicted material fields.}
    Two representative objects are compressed by the same prescribed rigid-plate
    motion using the ground-truth material field, the field predicted by
    VoMP~\citep{dagli2025vomp}, and the field predicted by ViWi. The filmstrips
    show the resulting deformation over time, while the plots on the right report
    trajectory deviation from the ground-truth-driven simulation, normalized by
    object size. ViWi remains substantially closer to the reference simulation in
    both examples.
    }
    \label{fig:squeeze}
\end{figure}

\paragraph{Quasi-Static Compression.}
To complement the transient impact test, we further evaluate the predicted
mechanical-property fields under global displacement-controlled compression. A rigid
plate descends onto each object until a prescribed compression of \(30\%\) of the
object's height is reached. As in the impact experiment, the scenario is simulated
using the ground-truth, VoMP, and ViWi mechanical-property fields under the same
initial state and reduced basis. The same prescribed plate motion is applied to
all three rollouts, and a common global stiffness scale is applied identically
across them, so that the comparison isolates differences in the predicted spatial
mechanical-property fields.

Figure~\ref{fig:squeeze} shows two representative compression examples. For the
bowl, the VoMP-driven simulation reaches a mean trajectory deviation of
\(0.64\%\) and a peak deviation of \(1.83\%\) of object size, whereas the
ViWi-driven simulation reduces these values to \(0.25\%\) and \(0.84\%\),
respectively. For the wheel-and-stand example, the corresponding deviations
decrease from \(0.32\%/1.65\%\) with VoMP to \(0.03\%/0.13\%\) with ViWi.
ViWi also more closely recovers the spatial stiffness structure, including the
softer support beneath the stiffer wheel.

Together, the impact and compression experiments demonstrate that the
mechanical-property fields predicted by ViWi produce physical behavior that more
closely follows the ground-truth-driven simulation under both localized dynamic
loading and global quasi-static deformation.

\begin{figure}[t]
    \centering
    \includegraphics[width=\textwidth]{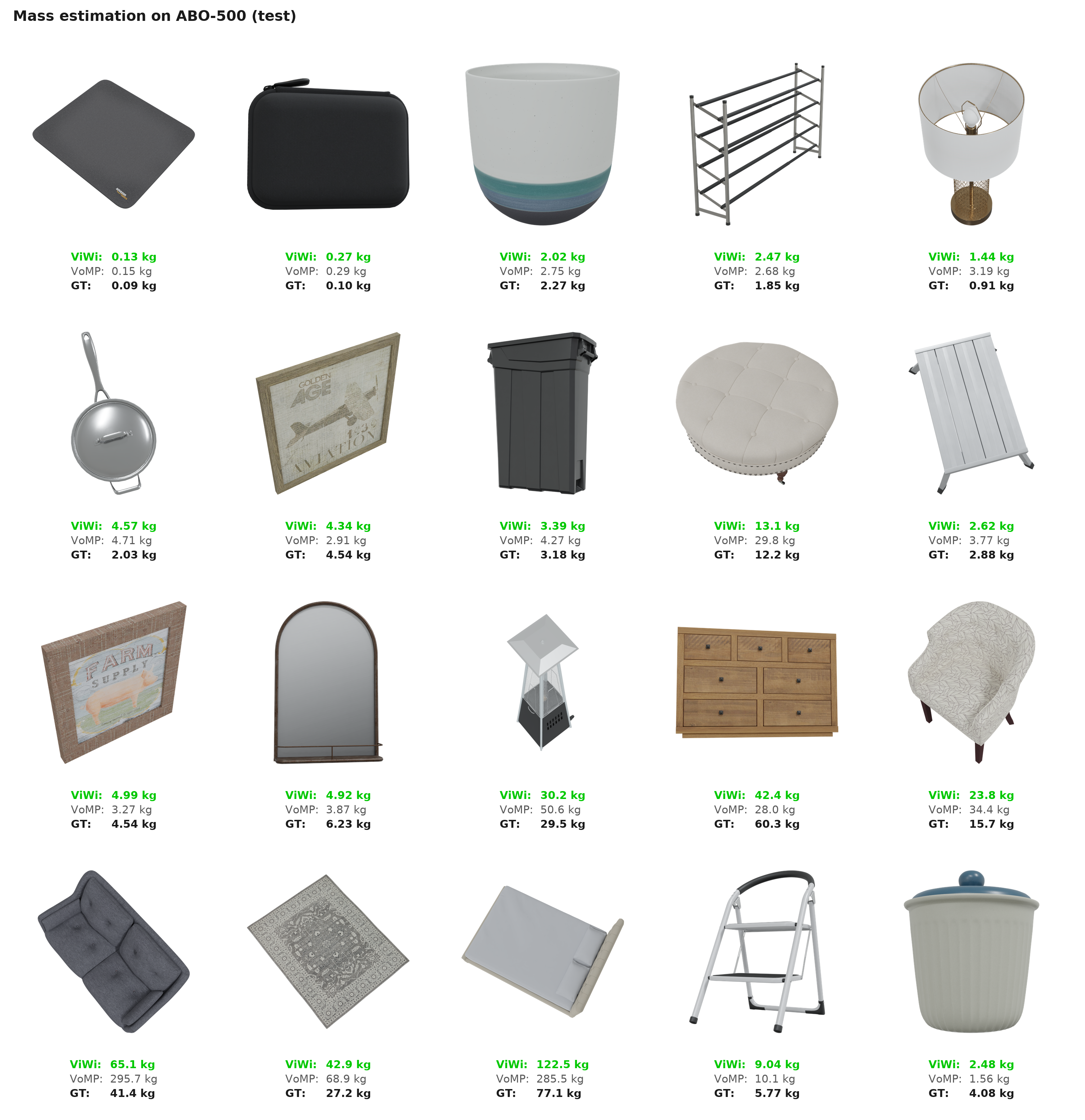}
    \caption{
    {Qualitative mass estimation on ABO-500.}
    Representative real objects from the ABO-500 test set are shown together
    with their ground-truth mass (GT) and the mass estimates obtained from
    VoMP~\citep{dagli2025vomp} and ViWi\(^{\dagger}\).
    ViWi\(^{\dagger}\) denotes the vision-only material-slot variant used for
    ABO-500 evaluation. Predicted masses are obtained from the estimated
    volumetric density fields.
    }
    \label{fig:abo_mass_qualitative}
\end{figure}

\paragraph{Qualitative Mass Estimation on ABO-500.}
Figure~\ref{fig:abo_mass_qualitative} provides qualitative examples of mass
estimation on the ABO-500 test set. For each real object, we report the
ground-truth mass together with the estimates obtained from VoMP and
ViWi\(^{\dagger}\), our vision-only material-slot variant. Across objects with
different geometries, scales, and appearances, ViWi\(^{\dagger}\) often produces
mass estimates closer to the ground truth than VoMP. These examples complement
the quantitative results in Table~\ref{tab:abo500_mass} and illustrate the
generalization of the predicted volumetric density fields from synthetic training
data to real-world objects.

\begin{figure}[t]
    \centering
    \includegraphics[width=\textwidth]{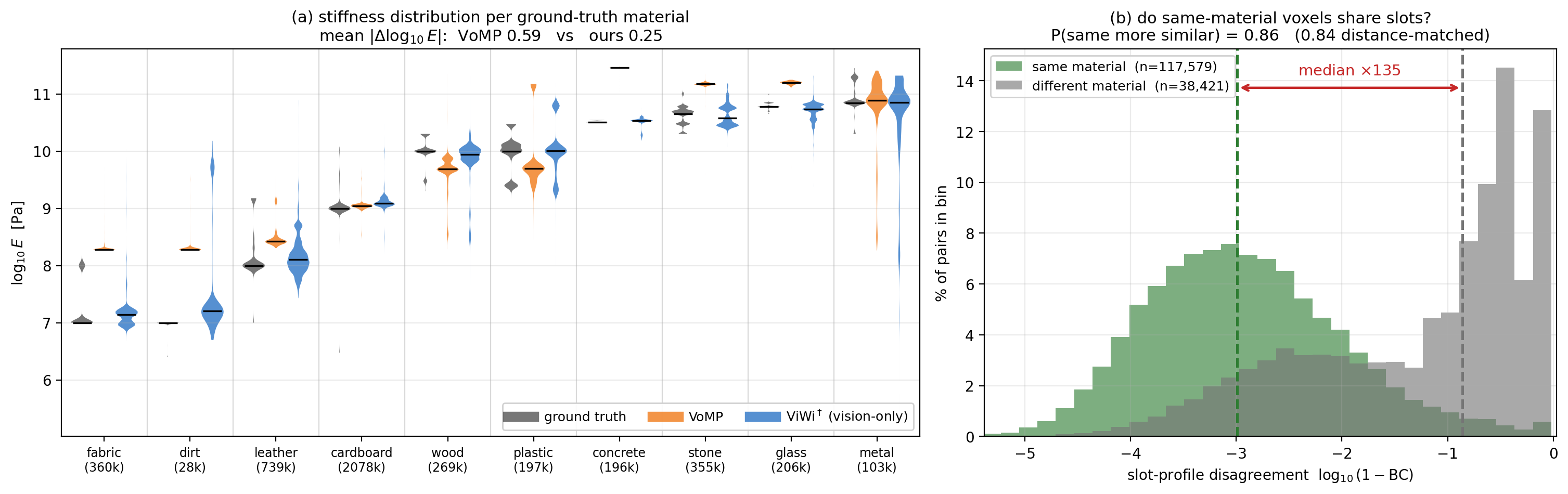}
\caption{
{Material consistency of the learned material slots.}
\textbf{(a)} Distribution of \(\log_{10}E\) per ground-truth material category.
ViWi$^{\dagger}$ tracks the ground truth more closely than the voxel-wise VoMP
baseline (mean \(|\Delta\log_{10}E|\): \(0.59\!\to\!0.25\)).
\textbf{(b)} Slot-assignment disagreement \(1-\mathrm{BC}\) for voxel pairs within
the same object: same-material pairs share slots far more than different-material
pairs (median \(\times135\)), and this survives matching on voxel distance
(\(0.86\!\to\!0.84\)), so slots capture material rather than proximity.
}
    \label{fig:slot_analysis}
\end{figure}

\paragraph{What the slots learn.}
Figure~\ref{fig:slot_analysis} looks inside the model in the vision-only setting, where
the slot head is the only difference from VoMP: both use the same frozen encoder
and the same MatVAE decoder, and neither uses RF.

Panel~(a) groups test voxels by their ground-truth material and plots the
predicted stiffness of each group. VoMP pushes soft and very stiff materials
toward the middle of the training range, while ViWi\(^{\dagger}\) keeps them
apart and predicts a tighter range within each material. Averaged over the ten
materials with at least \(5{,}000\) test voxels, the error in \(\log_{10}E\)
drops from \(0.59\) to \(0.25\); the largest gains are on fabric
(\(1.08\!\to\!0.27\)), dirt (\(1.34\!\to\!0.58\)) and concrete
(\(0.94\!\to\!0.05\)), and VoMP stays slightly ahead on cardboard and metal.

Panel~(b) asks whether the slots group voxels by material, even though they never
see a material label. Each voxel has an attention profile over the \(K\) slots,
which is a distribution, so we compare two voxels with the Bhattacharyya
coefficient \(\mathrm{BC}(p,q)=\sum_k\sqrt{p_kq_k}\), the standard overlap
between two distributions: \(\mathrm{BC}=1\) when the two voxels use exactly the
same slots and \(0\) when they share none. We plot the disagreement
\(1-\mathrm{BC}\), so small values mean two voxels were assigned to the same
slots. For each of the \(78\) test objects made of more than one material, we
take pairs of voxels from that object and split the pairs by whether the two
voxels have the same ground-truth material. Voxels of the same material are assigned to almost the same slots, while voxels of
different materials are not: the median disagreement is \(1.0\times10^{-3}\) for
same-material pairs and \(1.38\times10^{-1}\) for different-material pairs, a
factor of \(135\). Comparing one random pair of each type, the same-material pair
is the more similar one \(86\%\) of the time. This is not an artifact of nearby
voxels being smoothed together: repeating the comparison using only pairs that are
equally far apart still gives \(84\%\).

\end{document}